\pdfoutput=1

\documentclass[11pt]{article}

\usepackage{EMNLP2022}

\usepackage{times}
\usepackage{latexsym}

\usepackage{microtype}
\usepackage{graphicx}
\usepackage{subfigure}
\usepackage{booktabs} 

\usepackage{hyperref}

\usepackage{amsmath,amsfonts,bm}
\usepackage{xspace}
\usepackage{ulem}
\usepackage{wrapfig}
\newcommand{\jaccard}{$jaccard$}
\usepackage{pifont} 

\newcommand{\pmacro}{$p_\text{macro}$}
\newcommand{\pmicro}{$p_\text{micro}$}
\newcommand{\rmacro}{$r_\text{macro}$}
\newcommand{\rmicro}{$r_\text{micro}$}
\newcommand{\fmacro}{$F_\text{macro}$}
\newcommand{\fmicro}{$F_\text{micro}$}
\newcommand{\sts}{\textsc{seq2seq}\xspace}
\newcommand{\kset}{\ry_1, \ry_2, \ldots, \ry_k}
\newcommand{\vae}{\textsc{vae}\xspace}

\newcommand{\ours}{\textsc{setaug}\xspace}
\newcommand{\rand}{\textsc{random}\xspace}
\newcommand{\ar}{\textsc{ar}\xspace}
\newcommand{\nlp}{\textsc{nlp}\xspace}
\newcommand{\eg}{e.g.,\xspace}
\newcommand{\ie}{i.e.,\xspace}
\newcommand{\bart}{\textsc{bart}\xspace}
\newcommand{\tf}{\textsc{t5-11b}\xspace}
\newcommand{\bert}{\textsc{bert}\xspace}

\newcommand{\gpth}{\textsc{gpt3-175b}\xspace}
\newcommand{\lstm}{\textsc{lstm}\xspace}
\newcommand{\baseline}{\sts}
\newcommand{\elbo}{\textsc{elbo}\xspace}

\newcommand{\bsearch}{\textsc{set search}\xspace}
\newcommand{\card}{\textsc{card}\xspace}
\newcommand{\clsbaseline}{\textsc{multi-label}\xspace}
\newcommand{\clsbaselinewithk}{\textsc{multi-label-k*}\xspace}
\newcommand{\pre}{$p$\xspace}
\newcommand{\rec}{$r$\xspace}
\newcommand{\fs}{$F1$-\xspace}
\newcommand{\fst}{$F$\xspace}

\newcommand{\emoclass}{\textsc{go-emo}\xspace}
\newcommand{\openent}{\textsc{openent}\xspace}
\newcommand{\keygen}{\textsc{keygen}\xspace}

\newcommand{\reuters}{\textsc{reuters}\xspace}
\newcommand{\sample}{\textsc{tsample}\xspace}

\newcommand{\squishlist}{
  \begin{list}{$\bullet$}
    { \setlength{\itemsep}{0pt}      \setlength{\parsep}{3pt}
      \setlength{\topsep}{3pt}       \setlength{\partopsep}{0pt}
      \setlength{\leftmargin}{1.5em} \setlength{\labelwidth}{1em}
      \setlength{\labelsep}{0.5em} } }
\newcommand{\reallysquishlist}{
  \begin{list}{$\bullet$}
    { \setlength{\itemsep}{0pt}    \setlength{\parsep}{0pt}
      \setlength{\topsep}{0pt}     \setlength{\partopsep}{0pt}
      \setlength{\leftmargin}{0.2em} \setlength{\labelwidth}{0.2em}
      \setlength{\labelsep}{0.2em} } }

 \newcommand{\squishend}{
     \end{list} 
 }

\newcommand{\yy}[1]{}

\newcommand{\TODO}[1]{}

\newcommand{\bigo}[1]{\mathcal{O}(#1)}

\def\Figref#1{Figure~\ref{#1}}

\def\Secref#1{Section~\ref{#1}}

\def\eqref#1{equation~\ref{#1}}
\def\Eqref#1{Equation~\ref{#1}}

\def\1{\bm{1}}
\newcommand{\train}{\mathcal{D}}

\def\ry{{\textnormal{y}}}

\def\vx{{\bm{x}}}
\def\vy{{\bm{y}}}
\def\vz{{\bm{z}}}

\def\evy{{y}}

\def\mZ{{\bm{Z}}}

\DeclareMathAlphabet{\mathsfit}{\encodingdefault}{\sfdefault}{m}{sl}
\SetMathAlphabet{\mathsfit}{bold}{\encodingdefault}{\sfdefault}{bx}{n}

\newcommand{\expval}[2]{\mathbb{E}_{#1}\left[#2\right]}

\def\sY{{\mathbb{Y}}}

\newcommand{\pcondest}[2]{\hat{p}(#1 \mid #2)}
\newcommand{\pcond}[2]{p(#1 \mid #2)}
\newcommand{\pcondth}[2]{p_{\theta}(#1 \mid #2)}

\newcommand{\Ls}{\mathcal{L}}

\newcommand{\ind}{\perp \!\!\! \perp}

\usepackage{algorithm}
\usepackage{algorithmic}
\usepackage{wrapfig,lipsum,booktabs}
\usepackage{hyperref}
\usepackage{url}
\usepackage{graphicx}
\usepackage{multirow}
\usepackage{tikz}
\usepackage{color}
\usepackage{xspace}
\usetikzlibrary{bayesnet}
\usepackage{graphicx}
\usepackage{color}
\usepackage{amsmath}
\usepackage{floatflt}
\usepackage{soul}
\usepackage{xspace}
\usepackage{subfigure}
\usepackage{capt-of}

\usepackage{amsmath}
\usepackage{amssymb}
\usepackage{mathtools}
\usepackage{amsthm}

\usepackage[capitalize,noabbrev]{cleveref}

\theoremstyle{plain}
\newtheorem{theorem}{Theorem}[section]

\newtheorem{lemma}[theorem]{Lemma}

\theoremstyle{definition}

\theoremstyle{remark}

\usepackage[T1]{fontenc}

\usepackage[utf8]{inputenc}

\usepackage{microtype}

\title{Conditional set generation using \sts models}

\author{Aman Madaan, Dheeraj Rajagopal,  \textbf{Niket Tandon$^\dagger$},\\ \textbf{Yiming Yang}, 
\textbf{Antoine Bosselut$^\ddagger$} \\
  Language Technologies Institute, Carnegie Mellon University, Pittsburgh, PA, USA \\ 
  $^\dagger$ Allen Institute for Artificial Intelligence, Seattle, WA, USA \\ 
  $^\ddagger$ EPFL, Switzerland \\
  \texttt{amadaan@cs.cmu.edu}}

\begin{document}
\maketitle

\begin{abstract}
Conditional set generation learns a mapping from an input sequence of tokens to a set.
Several \nlp tasks, such as entity typing and dialogue emotion tagging, are instances of set generation. 
\sts models, a popular choice for set generation, treat a set as a sequence and do not fully leverage its key properties, namely order-invariance and cardinality.
We propose a novel algorithm for effectively sampling informative orders over the combinatorial space of label orders.
We jointly model the set cardinality and output by prepending the set size and taking advantage of the autoregressive factorization 
used by \sts models.
Our method is a model-independent data augmentation approach that endows any \sts model with the signals of order-invariance and cardinality.
Training a \sts model on this augmented data~(without any additional annotations) gets an average relative improvement of 20\% on four benchmark datasets across various models: \bart-base, \tf, and \gpth.\footnote{Code to use \ours available at: \url{https://setgen.structgen.com}}
\end{abstract}

\newcommand{\comet}{\textsc{comet}}
\newcommand{\bleu}{\textsc{bleu}}
\newcommand{\rouge}{\textsc{rouge}}
\newcommand{\emnlpadd}[1]{#1}
\newcommand{\emnlpcr}[1]{#1}

\section{Introduction}

Conditional set generation is the task of modeling the distribution of an output set given an input sequence of tokens~\citep{kosiorek2020conditional}.
Several \nlp tasks are instances of set generation, including open-entity typing~\citep{choi2018ultra,dai2021ultra}, fine-grained emotion classification~\citep{demszky2020goemotions}, and \emnlpadd{keyphrase generation~\citep{meng2017deep,yuan2018one,ye2021one2set}}.
The recent successes of the pretraining-finetuning paradigm have encouraged a formulation of set generation 
as a \sts  generation task~\citep{vinyals2015order,yang2018sgm,meng2019doesordermatter,ju2020transformer}.

In this paper, we posit that modeling set generation as a vanilla \sts generation task is sub-optimal, because the \sts formulations do not explicitly account for two key properties of a set output: \textit{order-invariance} and \textit{cardinality}. 
Forgoing order-invariance, vanilla \sts generation 
treats a set as a sequence, assuming an arbitrary order between the elements it outputs. Similarly, the cardinality of sets is ignored, as the number of elements to be generated is typically not modeled.

Prior work has highlighted the importance of these two properties for set output through loss functions that encourage order invariance~\citep{ye2021one2set}, exhaustive search over the label space for finding an optimal order~\citep{qin2019adapting,rezatofighi2018deep,vinyals2015order}, and post-processing the output~\citep{chowdhury2016know2look}.
Despite the progress, several important gaps remain.
First, exhaustive search does not scale with large output spaces typically found in \nlp problems, thus stressing the need for an optimal sampling strategy for the labels. 
Second, cardinality is still not explicitly modeled in the \sts setting despite being an essential aspect for a set.
Finally, architectural modifications required for specialized set-generation techniques might not be viable for modern large-language models.

\begin{figure*}[!t]
\centering
  \includegraphics[scale=0.19]{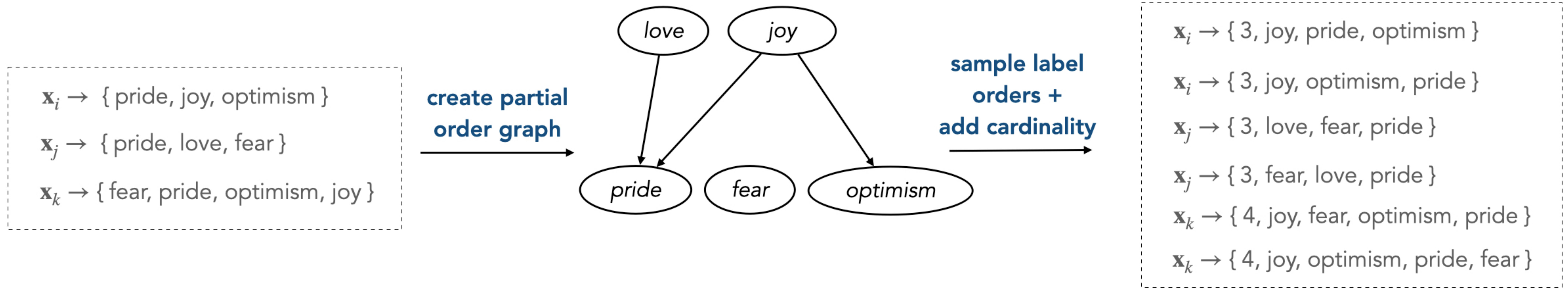}
\caption{
An illustrative task where given an input $\vx$, the output is a set of emotions. 
Our method first discovers a partial order graph~(middle) in which specific labels~(joy) come before more general labels~(pride).
Listing the specific labels first gives the model more clues about the rest of the set.
Topological samples from this partial order graph are label sequences that can be efficiently generated  using \sts models.
The size of each set is also added as the first element for joint modeling of output with size.}
\label{fig:method}
\end{figure*}

We address these challenges with a novel data augmentation strategy.
Specifically, we take advantage of the auto-regressive factorization used by \sts models and (i) impose an \textit{informative} order over the label space, and (ii) explicitly model \textit{cardinality}. 
First, the label sets are converted to sequences using informative orders by grouping labels and leveraging their dependency structure. 
Our method \emnlpcr{induces} a partial order graph over \emnlpcr{label space} where the nodes are the labels, and the edges denote the conditional dependence relations.
\emnlpcr{This graph provides a natural way to obtain informative orders while reinforcing order-invariance.
Specifically, sequences obtained via topological traversals of this graph allow independent labels to appear at different locations in the sequence, while restricting order for dependent labels.}
Next, we jointly model a set with its cardinality by simply prepending the set size to the output sequence.  
This strategy aligns with the current trend of very large language models which do not lend themselves to architectural modifications but increasingly rely on the informativeness of the inputs~\citep{yang2020g,Liu2021PretrainPA}.

\Figref{fig:method} illustrates the key intuitions behind our method using sample task where given an input $\vx$ (say a conversation), the output is a set of emotions~($\sY$).
To see why certain orders might be more meaningful, consider a case where one of the emotions is \textit{joy}, which leads to a more general emotion of \textit{pride}.
After first generating \textit{joy}, the model can generate \textit{pride} with certainty~(\textit{joy} leads to \textit{pride} in all samples). In contrast, the reverse order (generating \textit{pride} first) still leaves room for multiple possible emotions~(joy and love).
The order $[\text{joy}, \text{pride}]$ is thus more informative than $[\text{pride}, \text{joy}]$.
The cardinality of a set can also be helpful.
In our example, joy contains two sub-emotions, and love contains one.
A model that first predicts the number of sub-emotions can be more precise 
and avoid over-generation, a significant challenge with language generation models~\citep{welleck2019neural,fu2021theoretical}. 
We efficiently sample such informative orders from the combinatorial space of all possible orders and jointly model cardinality by leveraging the auto-regressive nature of \sts models.

\paragraph{Our contributions}
\begin{enumerate}

\item[(i)] We show an efficient way to model sequence-to-set prediction as a \sts task by jointly modeling the cardinality and augmenting the training data with informative sequences using our novel \ours data augmentation approach.~(\S \ref{sec:tsample}, \ref{sec:modelingcardinality}).
\item[(ii)] We theoretically ground our approach: treating the order as a latent variable, we show that our method serves as a better proposal distribution in a variational inference framework.~(\S\ref{par:vi})
\item[(iii)] With our approach, \sts models of different sizes achieve a $\sim$20\% relative improvement on four real-world tasks, with no additional annotations or architecture changes.~(\S\ref{sec:experiments}).
\end{enumerate}
\section{Task}
\label{sec:background}
We are given a corpus $\train = \{(\vx_t, \sY_t)\}_{t=1}^{m}$ where $\vx_t$ is a sequence of tokens and $\sY_t=\{\ry_1, \ry_2, \ldots, \ry_k\}$ is a set.
For example, in multi-label fine-grained sentiment classification, $\vx_t$ is a paragraph, and  $\sY_t$ is a set of sentiments expressed by the paragraph.
We use $\ry_i$ to denote an output symbol, $[\ry_i, \ry_j, \ry_k]$ to denote an ordered sequence of symbols and $\{\ry_i, \ry_j, \ry_k\}$ to denote a set.

\subsection{Set generation using \sts model}

\paragraph{Task} Given a corpus $\{(\vx_t, \sY_t)\}_{t=1}^{m}$, the task of conditional set generation is to efficiently estimate $\pcond{\sY_t}{\vx_t}$.
\sts models factorize $\pcond{\sY_t}{\vx_t}$ autoregressively~(\ar) using the chain rule: 

{\small
$$
\begin{aligned}
    \pcond{\sY_t}{\vx_t} &= \pcond{\ry_1, \ry_2, \ldots, \ry_k}{\vx_t} \nonumber \\ 
    &=\pcond{\ry_1}{\vx_t}\prod_{j=2}^{k}\pcond{\ry_j}{\vx_i, \ry_1\ldots\ry_{j-1}}
\end{aligned}
$$
}

where the order $\sY_t = [\ry_1, \ry_2, \ldots, \ry_k]$ factorizes the joint distribution using chain rule.
In theory, any of the $k!$ orders can be used to factorize the same joint distribution.
In practice, the choice of order is important. 
For instance, \citet{vinyals2015order} show that output order affects language modeling performance when using \lstm based \sts models for set generation.

Consider an example input-output pair $(\vx_t, \sY_t = \{\ry_1, \ry_2\})$.
By chain rule, we have the following equivalent factorizations of this sequence:
$\pcond{\sY_t}{\vx_t    } = \pcond{\ry_1}{\vx}\pcond{\ry_2}{\vx, \ry_1} = \pcond{\ry_2}{\vx}\pcond{\ry_1}{\vx, \ry_2}$.
However, order-invariance is only guaranteed with \textit{true} conditional probabilities, whereas the conditional probabilities used to factorize a sequence are \textit{estimated} by a model from a corpus.
Thus, dependening on the order, the sequence factorizes as either $\pcondest{\ry_1}{\vx}\pcondest{\ry_2}{\vx, \ry_1}$ or $\pcondest{\ry_2}{\vx}\pcondest{\ry_1}{\vx, \ry_2}$, which are not necessarily equivalent. 
\emnlpcr{Further, one of these two factorizations may be better represented in the training data, and thus lead to better samples. For instance, if the training data always contains $\ry_1, \ry_2$ in the order [$\ry_1, \ry_2$], $\pcondest{\ry_2}{\vx}\pcondest{\ry_1}{\vx, \ry_2}$ will be $\sim0$.}

\emnlpcr{Order will also be immaterial if the labels are conditionally independent given the input~(\Secref{sec:ordershouldnotmatter}). However, this assumption is often not valid in practice, especially for \nlp, where labels typically share a semantic relationship.}

\section{Method}

\label{sec:method}
This section expands on two critical components of our system, \ours. 
\Secref{sec:tsample} presents \sample, a method to create informative orders over sets tractably. \Secref{sec:modelingcardinality} presents our method for jointly modeling cardinality and set output.

\subsection{\sample: Adding informative orders for set output}
\label{sec:tsample}

As discussed in~\Secref{sec:background}, \sts formulation requires the output to be in a sequence.
Prior work \citep{vinyals2015order,rezatofighi2018deep,chen2021order} has noted that listing the output in orders that have the highest conditional likelihood given the input is an optimal choice.
Unlike these methods, we sidestep exhaustive searching during training using our proposed approach \sample.


Our core insight is that knowing the optimal order between pairs of symbols in the output drastically reduces the possible number of permutations. 
We thus impose pairwise order constraints for labels.
Specifically, given an output set $\sY_t = \kset$, if $\ry_i, \ry_j$ are independent, they can be added in an arbitrary order.
Otherwise, an order constraint is added to the order between $\ry_i, \ry_j$.

\paragraph{Learning pairwise constraints} We estimate the dependence between elements $\ry_i, \ry_j$ using pointwise mutual information: $\texttt{pmi}(\ry_i, \ry_j) = \log p(\ry_i, \ry_j) / p(\ry_i)p(\ry_j)$.
Here, $\texttt{pmi}(\ry_i, \ry_j) > 0$ indicates that the labels $\ry_i, \ry_j$ co-occur more than would be expected under the conditions of independence~\citep{wettler-rapp-1993-computation}.
We use $\texttt{pmi}(\ry_i, \ry_j) > \alpha$ to filter our such pairs of dependent pairs, and perform another check to determine if the order between them should be fixed.
For each dependent pair $\ry_i, \ry_j$, the order is constrained to be $[\ry_i, \ry_j]$~($\ry_j$ should come after $\ry_i$) if $\log \pcond{\ry_j}{\ry_i} - \log \pcond{\ry_i}{\ry_j} > \beta $, and $[\ry_j, \ry_i]$ otherwise. 
Intuitively, $\log \pcond{\ry_j}{\ry_i} - \log \pcond{\ry_i}{\ry_j} > \beta$ implies that knowledge that a set contains $\ry_i$, increases the probability of $\ry_j$ being present.
Thus, \emnlpcr{in an autoregressive setting}, fixing the order to $[\ry_i, \ry_j]$ will be more efficient for generating a set with $\{\ry_i, \ry_j\}$.

\begin{figure}[]
    \centering
\includegraphics[width=0.48\textwidth]{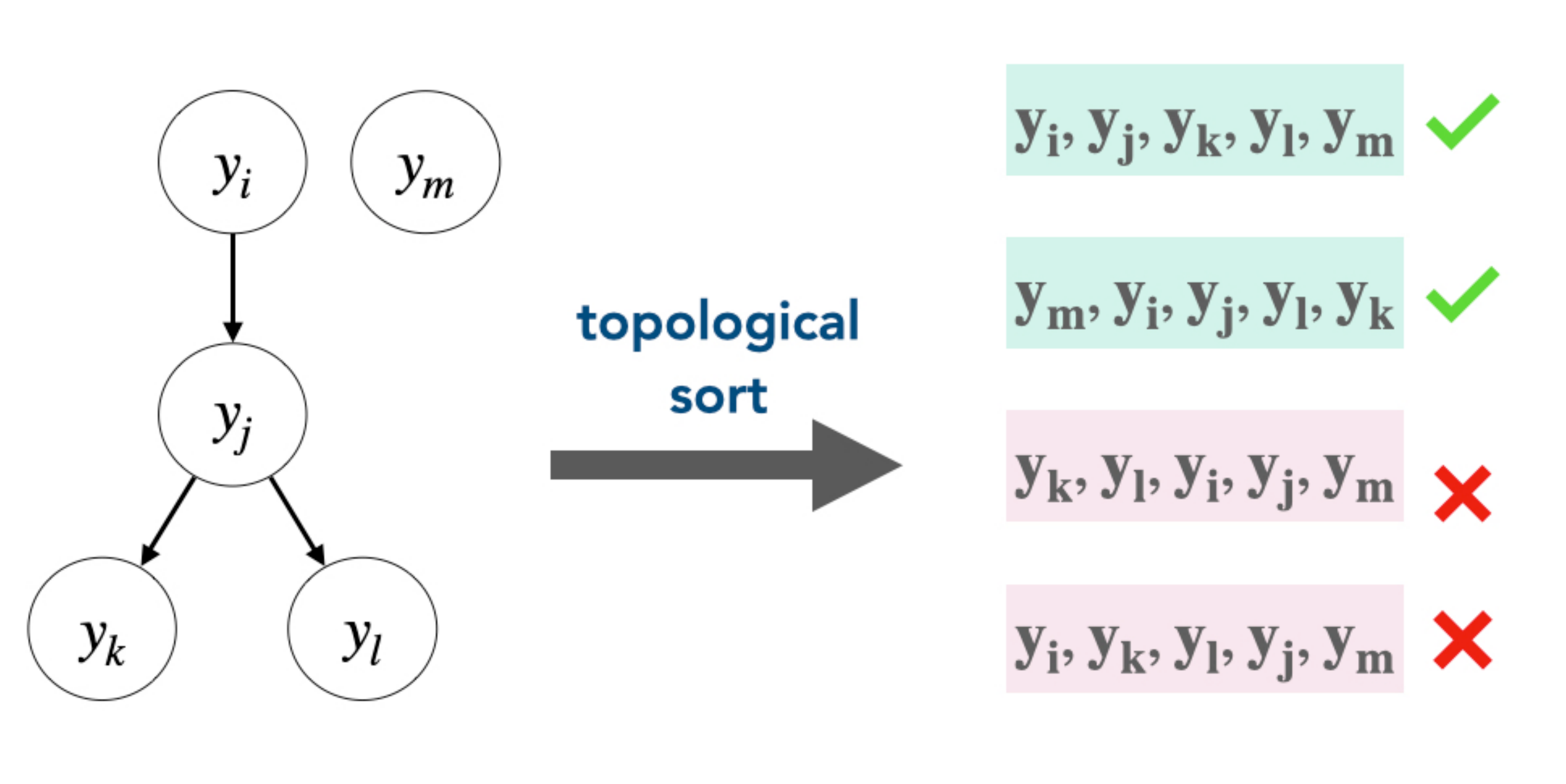}
\captionof{figure}{Our sampling method \sample first builds a graph $G_t$ over the set $\sY_t$, and then samples orders from $G_t$ using topological sort (\texttt{topo\_sort}). The topological sorting rejects samples that do not follow the conditional probability constraints.}
\label{fig:penope}
\end{figure}

\paragraph{Generating samples} To systematically create sequences that satisfy these constraints, we construct a topological graph $G_t$ where each node is a label $\vy_i \in \sY_t$, and the edges are determined using the \texttt{pmi} and the conditional probabilities as outlined above~(Algorithm \ref{alg:topographcreation}).
The required permutations can then be generated as topological traversals $G_t$~(\Figref{fig:penope}). 
We begin the traversal from a different starting node to generate diverse samples. 
We call this method \sample.
Our method of generating graphs avoids cycles by design~(proof in \ref{sec:nocycles}), and thus topological sort remains well-defined.
Later, we show that \sample can be interpreted as a proposal distribution in variational inference framework, which distributes the mass uniformly over informative orders constrained by the graph. 

\paragraph{Do pairwise constraints hold for longer sequences?} 
While \sample uses pairwise~(and not higher-order) constraints for ordering variables, we note that the pairwise checks remain relevant with extra variables.
First, dependence between pair of variables is retained in joint distributions involving more variables~($\ry_i \not \ind \ry_j \implies \ry_i \not \ind \ry_j, \vy_k$) for some $\vy_k \in \sY$~(Appendix~\ref{sec:higherorderdeps}).
Further, if $\ry_i, \ry_j \ind \vy_k$, then it can be shown that $\pcond{\ry_i}{\ry_j} > \pcond{\ry_j}{\ry_i} \implies \pcond{\ry_i}{\ry_j, \vy_k} > \pcond{\ry_j}{\ry_i, \vy_k}$~(Appendix~\ref{sec:conditioningretainsprobs}).
The first property shows that the pairwise dependencies hold in the presence of other set elements.
The second property shows that an informative order continues to be informative when additional independent symbols are added. 
Thus, 
using pairwise dependencies between the set elements is still effective.
Using higher-order dependencies might be suboptimal for practical reasons: higher-order dependencies~(or including $\vx_t$) might not be accurately discovered due to sparsity, and thus cause spurious orders.

Finally, we note that if all the labels are independent, then the order is guaranteed not to matter ~(Lemma~\ref{sec:ordershouldnotmatter}, also shown empirically in Appendix~\ref{sec:simulationcomplete}).
Thus, our method will only be useful when labels have some degree of dependence.

\paragraph{Complexity analysis}
Let $\sY$ be the label space, $(\vx_t, \sY_t)$ be a particular training example, $N$ be the size of the training set, and $c$ be the maximum number of elements for any set $\sY_t$ in the input.
Our method requires three steps: i) iterating over training data to learn conditional probabilities and pmi, and ii) given a $\sY_t$, building the graph $G_t$ ~(Algorithm~\ref{alg:topographcreation}), and iii) doing topological traversals over $G_t$ to create samples for $(\vx_t, \sY_t)$.

The time complexity of the first operation is $\bigo{Nc^2}$: for each element of the training set, the pairwise count for each pair $\ry_i, \ry_j$ and unigram count for each $\ry_i$ is calculated.
The pairwise counts can be used for calculating joint probabilities.
In principle, we need $\bigo{|\sY|^2}$ space for storing the joint probabilities.
In practice, only a small fraction of the combinations will appear $|\sY|^2$ in the corpus.

Given a set $\sY_t$ and the conditional probabilities, the graph $G_t$ is created in $\bigo{c^2}$ time.
Then, generating $k$ samples from $G_t$ requires a topological sort, for $\bigo{kc}$~(or $\bigo{c}$ per traversal).
For training data of size $N$, the total time complexity is $\bigo{Nck}$.
The entire process of building the joint counts and creating graphs and samples takes less than five minutes for all the datasets on an 80-core Intel Xeon Gold 6230 CPU.

\begin{algorithm}[!t]
\small
\textbf{Input}: Set $\sY_t$, number of permutations $n$\\
\textbf{Parameter}: $\alpha, \beta$\\
\textbf{Output}: $n$ topological sorts over $G_t(V, E)$ \\
\begin{algorithmic}[1] 
\STATE Let $V=\sY_t, E=\emptyset$.
\FOR{$\ry_i, \ry_j \in \sY_t$}
\IF {$pmi(\ry_i, \ry_j) > \alpha$; $\lg \pcond{\ry_i}{\ry_j} - \lg \pcond{\ry_j}{\ry_i} > \beta$ }
\STATE $E = E \cup \ry_j \rightarrow \ry_i$ 
\ENDIF
\ENDFOR
\STATE \textbf{return} $\texttt{topo\_sort}(G_t(V, E), n)$
\end{algorithmic}
\caption{Generating permutations for $\sY_t$}
\label{alg:topographcreation}
\end{algorithm}

\paragraph{Interpreting \sample as a proposal distribution over orders}
\label{par:vi}
We show that our method of augmenting permutations to the training data can be interpreted as an instance of variational inference with the order as a latent variable, and \sample as an instance of a richer proposal distribution.

Let $\pi_j$ be the $j^{th}$ order over $\sY_t$ (out of $|\sY_t|!$ possible orders $\Pi$), and $\pi_j(\sY_t)$ be the sequence of elements in $\sY_t$ arranged with order $\pi_j$.
Treating $\pi$ as a latent random variable, the output distribution can then be recovered by marginalizing over all possible orders $\Pi$: 
\begin{align*}
    \log \pcondth{\sY_t}{\vx_t} &= \log \sum_{\pi_{z} \in \Pi} \pcondth{\pi_{z}(\sY_t)}{\vx_t}
\end{align*}
where $p_{\theta}$ is the \sts conditional generation model.
While summing over $\Pi$ is intractable, standard techniques from the variational inference framework allow us to write a lower bound~(\elbo) on the actual likelihood:

{
\small
$$
\begin{aligned}
    \log \pcondth{\sY_t}{\vx_t}\nonumber &= \log \sum_{\pi_\vz \in \Pi}\pcondth{\pi_{z}(\sY_t)}{\vx_t} \nonumber\\
    &\geq  \underbrace{\expval{q_{\phi}(\pi_\vz)}{\log\frac{ \pcondth{\pi_\vz(\sY_t)}{\vx_t}}{q_{\phi}(\pi_\vz)}}}_{\textsc{elbo}}\nonumber = \Ls(\theta, \phi) \\ \nonumber
\end{aligned}
$$
}

In practice, the optimization procedure draws $k$ samples from the proposal distribution $q$ to optimize a weighted \elbo ~\citep{burda2016importance,domke2018importance}.
Crucially, $q$ can be fixed~(\eg to uniform distribution over the orders), and in such cases only $\theta$ are learned~(Appendix~\ref{sec:appendixvae}). 
\emnlpcr{The data augmentation approach used for \textsc{xl-net} \citep{yang2019xlnet} can be interpreted with this framework.
In their case, the proposal distribution $q$ is fixed to a uniform distribution for generating orders over tokens.
The variational interpretation also indicates that it might be possible to improve language modeling by using a different, more informative $q$. 
Investigating such proposal distribution for language modeling is an interesting future work.}

\sample can thus be seen as a particular proposal distribution that assigns all the support to the topological ordering over the label dependence graphs.
We experiment with sampling from a uniform distribution over the samples~(referred to as \rand experiments in our baseline setup).
The idea of using an informative proposal distribution over space of structures to do variational inference has also been used in the context of grammar induction~\citep{dyer2016recurrent} and graph generation~\citep{jin2018junction,chen2021order}.
Our formulation is closest in spirit to~\citet{chen2021order}. 
However, the set of nodes to be ordered is already given in their graph generation setting. In contrast, we infer the order and the set elements jointly from the input.

\subsection{Modeling cardinality}
\label{sec:modelingcardinality}

Let $m=|\sY_t|$ be the cardinality (number of elements) in $\sY_t$.
Our goal is to jointly estimate $m$ and $\sY_t$ (\ie $\pcond{m, \sY_t}{\vx_t}$).
Additionally, we want the model to use the cardinality information for generating $\sY_t$.
We add the cardinality information at the beginning of the sequence, thus converting a sample $(\vx_t, \sY_t)$ to $(\vx_t, [|\sY_t|, \pi(\sY_t)])$, and then train our \sts model as usual from $\vx \rightarrow [|\sY_t|, \pi(\sY_t)]$.
Here $\pi$ is some ordering that converts the set $\sY_t$ to a sequence.
As \sts models use autoregressive factorization, listing the order information first ensures that the distribution $\pcond{[|\sY_t|, \pi(\sY_t)]}{\vx_t}$ factorizes as $\pcond{[|\sY_t|, \pi(\sY_t)]}{\vx_t} = \pcond{|\sY_t|}{\vx_t}\pcond{\pi(\sY_t)}{|\sY_t|, \vx_t}$.
Thus, the generation of $\sY_t$ is conditioned on the input and the cardinality $|\sY_t|$~(due to $\pcond{\pi(\sY_t)}{|\sY_t|, \vx_t}$ term).

\paragraph{Why should cardinality help?}
\TODO{comments from Ruohong on cardinality helping: think about it in terms of the conditional distribution of EOS given y}
Unlike models like deep sets~\citep{zhang2019deep}, \sts models are not restricted by the number of elements generated. 
However, adding cardinality information 
has two potential benefits: i) it can help avoid over-generation~\citep{welleck2019neural,fu2021theoretical}, and ii) unlike free-form text output, the distribution of the set output size~($\pcond{|\sY_t|}{\vx_t}$) might benefit the model to adhere to the set size constraint. 
Thus, information on the predicted size can be beneficial for the model to predict the elements to be generated.





\section{Experiments}
\label{sec:experiments}
\ours comprises: i) \sample, a way to generate informative orders to convert sets to sequences, and ii) \card: jointly modeling cardinality and the set output. This section answers two questions:  

\reallysquishlist
\item [RQ1:] \textbf{How well does \ours improve existing models?} Specifically, how well \ours can take an existing \sts model and improve it just using our data augmentation and joint cardinality prediction, without making any changes to the model architecture. We also measure if these performance improvements carry across diverse datasets, model classes, and inference settings.
\item [RQ2:] \textbf{Why does our approach improve performance?} We study the contributions of \sample and joint cardinality prediction~(\card), and analyze where \ours works or fails. 
\squishend

\subsection{Setup}

\paragraph{Tasks} 
We consider multi-label classification and keyphrase generation. These tasks represent set generation problems where the label space spans a set of fixed categories (multi-label classification) or free-form phrases (keyphrase generation).

\noindent 1. \textbf{Multi-label classification task}: We have three datasets of varying sizes and label space:
\squishlist
    \item Go-Emotions classification (\emoclass, \citet{demszky2020goemotions}): generate a set of emotions for a paragraph.
    \item Open Entity Typing (\openent, \citet{choi2018ultra}): assigning open types~(free-form phrases) to the tagged entities in the input text.
    \item Reuters-21578 (\reuters, \citet{lewis1997reuters}): labeling news article with the set of mentioned economic subjects.
\squishend

\noindent 2. \textbf{Keyphrase generation~(\keygen)}: We experiment with a popular keyphrase generation dataset, KP20K~\citep{meng2017deep} which involves generating keyphrases for a scientific paper abstract.

Table~\ref{tab:data} lists the dataset statistics and examples from each dataset are shown in Appendix~\ref{dataset:examples}. We treat all the problems as open-ended generation, and do not use any specialized pre-processing. For all the datasets, we filter out samples with a single label. For each training sample, we create $n$ permutations using \ours.

\paragraph{Baselines} 

We compare with two baselines:\\
\noindent i) \textbf{\clsbaseline}: As a non-\sts baseline, we train a multi-label classifier that makes independent predictions of the output labels. Encoder-only and encoder-decoder approaches can be adapted for \clsbaseline, and we experiment with \bart~(encoder-decoder) and \bert~(encoder-only).
This baseline represents a standard method for doing multi-label classification~(\eg \citet{demszky2020goemotions}).
During inference, top-k logits are returned as the predicted set. 
We search over $k=[1, 3, 5, 10, 50]$ and use $k$ that performs the best on the dev set. Table~\ref{tab:resultsv3withclassifieradditional} in Appendix~\ref{sec:additionalresultsappendix} shows precision, recall, and \fst scores at each-k.

\noindent ii) \textbf{\bsearch}: each training sample $(\vx, \{\ry_1, \ry_2, \ldots, \ry_k\})$ is converted into $k$ training examples $\{(\vx, \ry_i)\}_{i=1}^{k}$.
We fine-tune \bart-base to generate one training sample for input $\vx$.
During inference, we run beam-search with the maximum set size in the training data~(Table~\ref{tab:data}).
The unique elements generated by beam search are returned as the set output, a popular approach for one-to-many generation tasks~\citep{hwang2021comet}.

\ours can apply to \textit{any} \sts model. We show results with models of various capacity:
\reallysquishlist
\item [iii)] \bart-base~\citep{lewis2020bart} (110M), 
\item [iv)] \tf~\citep{raffel2020exploring} (11B), and 
\item [v)] \gpth~\citep{Brown2020GPT3} (175B). 
\squishend

\begin{table}[!t]
\setlength{\tabcolsep}{0.3em}
\small
\begin{tabular}{@{}lccc@{}}
\toprule
Task &
  \begin{tabular}[c]{@{}c@{}}Avg/min/max \\  labels per sample\end{tabular} &
  \begin{tabular}[c]{@{}c@{}}Unique\\ labels\end{tabular} &
  \begin{tabular}[c]{@{}c@{}} Train/test/dev \\ samples per split\end{tabular}\\ \midrule
  \emoclass &
  3.03/3/5 &
  28 &
  0.6k/0.1k/0.1k \\
\openent &
  5.4/2/18 &
  2519 &
  2k/2k/2k \\
 \reuters & 2.52/2/11 & 90 & 0.9k/0.4k/0.3k \\ 
 \keygen & 3.87/3/79 & 274k & 156k/2k/2k
 \\\bottomrule
\end{tabular}
\caption{Datasets used in our experiments.}
\label{tab:data}
\end{table}

\paragraph{Training} 
We augment $n=2$ permutations to the original data using \sample.
For all the results, we use three epochs and the same number of training samples (\ie input data for the baselines is oversampled).
This controls for models trained with augmented data improving only because of factors such as longer training time.
All the experiments were repeated for three different random seeds, and we report the averages.
We found from our experiments\footnote{We conduct a one-tailed proportion of samples test~\citep{johnson2000probability} to compare with the strongest baseline, and underscore all results that are significant with $p<0.0005$.
For Algorithm~\ref{alg:topographcreation}, we try $\alpha=\{0.5, 1, 1.5 \}$ and $\beta = \{\log_2(2), \log_2(3), \log_2(4) \}$, and use networkx implementation of topological sort~\citep{hagberg2008exploring}.
} that hyperparameter tuning over $\alpha, \beta$ did not affect the results in any significant way.
For all the experiments reported, we use $\alpha=1$ and $\beta=\log_2(3)$. 
We use a single GeForce RTX 2080 Ti for all our experiments on bart, and a single TPU for all experiments done with T5-11B.
For \gpth, we use the OpenAI completion engine~(davinci) API \citep{openai_api}.
Additional hyperparameter details in Appendix~\ref{sec:hyperparams}.
We use greedy sampling for all experiments. Our method remains effective across five different sampling techniques, incl. beam search, nucleus, top-k, and random sampling~(Table~\ref{tab:samplingtypesetoverlapappendix}, Appendix~\ref{sec:simulationcomplete}).

\subsection{\ours improves existing models}
Our method helps across a wide range of models (\bart, \tf, and \gpth) and tasks.

\subsubsection{Multi-label classification}  
Table~\ref{tab:mainresults} shows improvements across all datasets and models for the multi-label classification task~($\sim$20\% relative gains). For brevity, we list macro \fst score, and include detailed results including macro/micro precision, recall, \fst scores in Table~\ref{tab:mainresultsdetailed}~(Appendix~\ref{sec:additionalresultsappendix}). We attribute the comparatively lower performance of \bsearch baseline to two specific reasons - repeated generation of the same set of terms~(\eg \textit{person, business} for \openent) and generating elements not present in the test set (see~\Secref{sec:erroranalysis} for a detailed error analysis).
We see similar trends with \gpth~(\Secref{sec:gpt3}).

\begin{table}[!t]
\small
    \centering
    \begin{tabular}{lrrr} \toprule
                 & \emoclass    &  \openent &  \reuters \\ \midrule
  \bsearch~(\bart)       &  7.4 & 26.3 & 7.5 \\
 \clsbaseline~(\bart)   &  25.6  &  16.4 &  25.2  \\
  \clsbaseline~(\bert)   &  25.7 &  16.2 &  25.5
 \\ \midrule
\bart            &  23.4 & 44.6 & 15.6  \\
\bart + \ours            &  \textbf{30.0} & \emph{\textbf{53.5}} & \emph{\textbf{26.7}}  \\ \midrule
\tf            &  47.8 & 53.6 & 45.3  \\
\tf + \ours            &  \textbf{50.9} & \emph{\textbf{57.0}} & \emph{\textbf{48.5}}  \\ \bottomrule
    \end{tabular}
    \caption{\ours improves \sts models by $\sim$20\% relative \fs points, on three multilabel classification datasets. \bart and \tf are trained on the original datasets with a random order and no cardinality. ``+ \ours'' indicates augmented train data using \sample and cardinality is prepended to the output sequence.}
    \label{tab:mainresults}
\end{table}

\begin{table}[!b]
\centering
\small
\begin{tabular}{@{}lrrr@{}}
\toprule
 \citet{ye2021one2set} & \bart & \bart + \ours \\ \midrule
5.8           & 5.3        & 6.5                    \\ 
39.2           & 36.3       & 39.1                    \\ \bottomrule
\end{tabular}%
\caption{ \ours \emnlpadd{improves off-the-shelf \bart-base for keyphrase generation task}
}
\label{tab:keygenresults}
\end{table}

\subsubsection{Keyphrase generation}
\label{sec:keygensection}
To further motivate the utility of \sts models for set generation tasks, we experiment on KP-20k, which is an extreme multi-label classification dataset~\citep{meng2017deep} with label space spanning over 257k unique keyphrases.
Due to the large label space, training multi-class classification baselines is not computationally viable.
\emnlpadd{In this dataset, the input text is an abstract from a scientific paper. We use the splits used by~\citet{ye2021one2set}.}
For a fair comparison with \citet{ye2021one2set}, we use \bart-base for this experiment.
Table~\ref{tab:keygenresults} shows the results.
Similar to datasets with smaller label space, our method improves on vanilla \sts.

\emnlpadd{We want to emphasize that while specialized models for individual tasks might be possible, we aim to propose a general approach that shows that sampling informative orders can help efficient and general set-generation modeling.}


\subsubsection{Simulations}
Following prior work on studying deep network properties effectively via simulation~\citep{vinyals2015order,khandelwal2018sharp}, we design a simulation to study the effects of output order and cardinality on conditional set generation.
The simulation reveals several key properties of our methods.
We defer the details to Appendix~\ref{sec:simulationcomplete}, and mention some key findings here.
We find similar trends in simulated settings.
Specifically, our method is (i) ineffective when labels are independent, (ii) helpful even when position embeddings are disabled, and (iii) helps across a wide range of sampling types. 

\subsubsection{Few-shot prompting with \gpth}
\label{sec:gpt3}
We fine-tune the generation models using augmented data for both \bart and \tf. However, fine-tuning models at the scale of \gpth is prohibitively expensive. Thus such models are typically used in \textit{a few-shot prompting setup}.
In a few-shot prompting setup, $M$ ($\sim$10-100) input-output examples are selected as a prompt $p$. A new input $x$ is appended to the prompt $p$, and $p \| x$ is the input to \gpth.
Improving these prompts, sometimes referred to with an umbrella term prompt tuning~\citep{Liu2021PretrainPA}, is a popular and emergent area of \nlp.
Our approach is the only feasible candidate for such settings, as it does not involve changing the model or additional post-processing.
We apply our approach for tuning prompts for generating sets in few-shot settings.\footnote{We use the text-davinci-001 version of \gpth available via the OpenAI API: \url{https://beta.openai.com/}}
We focus on \emoclass and \openent tasks, as the relatively short input examples allow cost-effective experiments.
We randomly create a prompt with $M=24$ examples from the training set and run inference over the test set for each.
For each example in the prompt, we order the set of emotions using our ordering approach \sample and compare the results with random orderings.
Using \sample to arrange the labels outperforms random ordering for both \openent~(macro \fst 34 vs. 39.5 with ours, 15\% statistically significant relative improvement), and \emoclass ~(macro \fst 16.5 vs. 14.5, 14\% relative improvement).
This suggests that ordering  
helps performance in 
resource-constrained settings e.g., few-shot prompting.


\begin{figure}[!ht]
    \centering
\includegraphics[width=0.8\columnwidth]{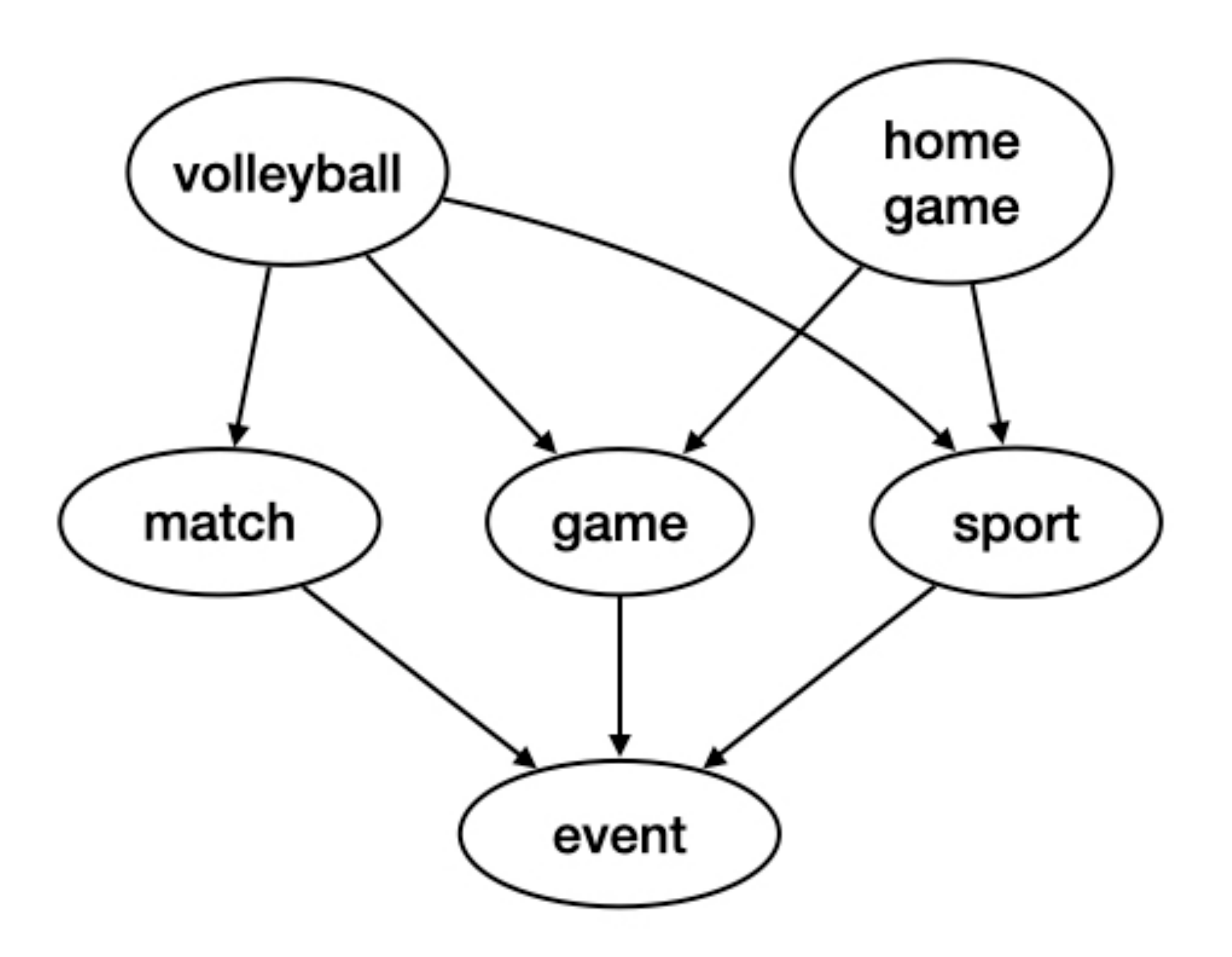}
\captionof{figure}{Label dependency discovered by \sample for \openent: specific entities~(\eg volleyball)
precede generic ones~(event). Appendix~\ref{sec:moreexamples} has more examples}
\label{fig:tsampletree}
\end{figure}

\subsection{Why does \ours improve performance?}
As mentioned in \Secref{sec:method}, our method of generating sets with \sts models consists of two components: i) a strategy for sampling informative orders over label space~(\sample), and ii) jointly generating cardinality of the output~(\card). 
This section studies the individual contributions of these components in order to answer RQ2.

\begin{figure}[!ht]
    \centering
\includegraphics[width=0.9\columnwidth]{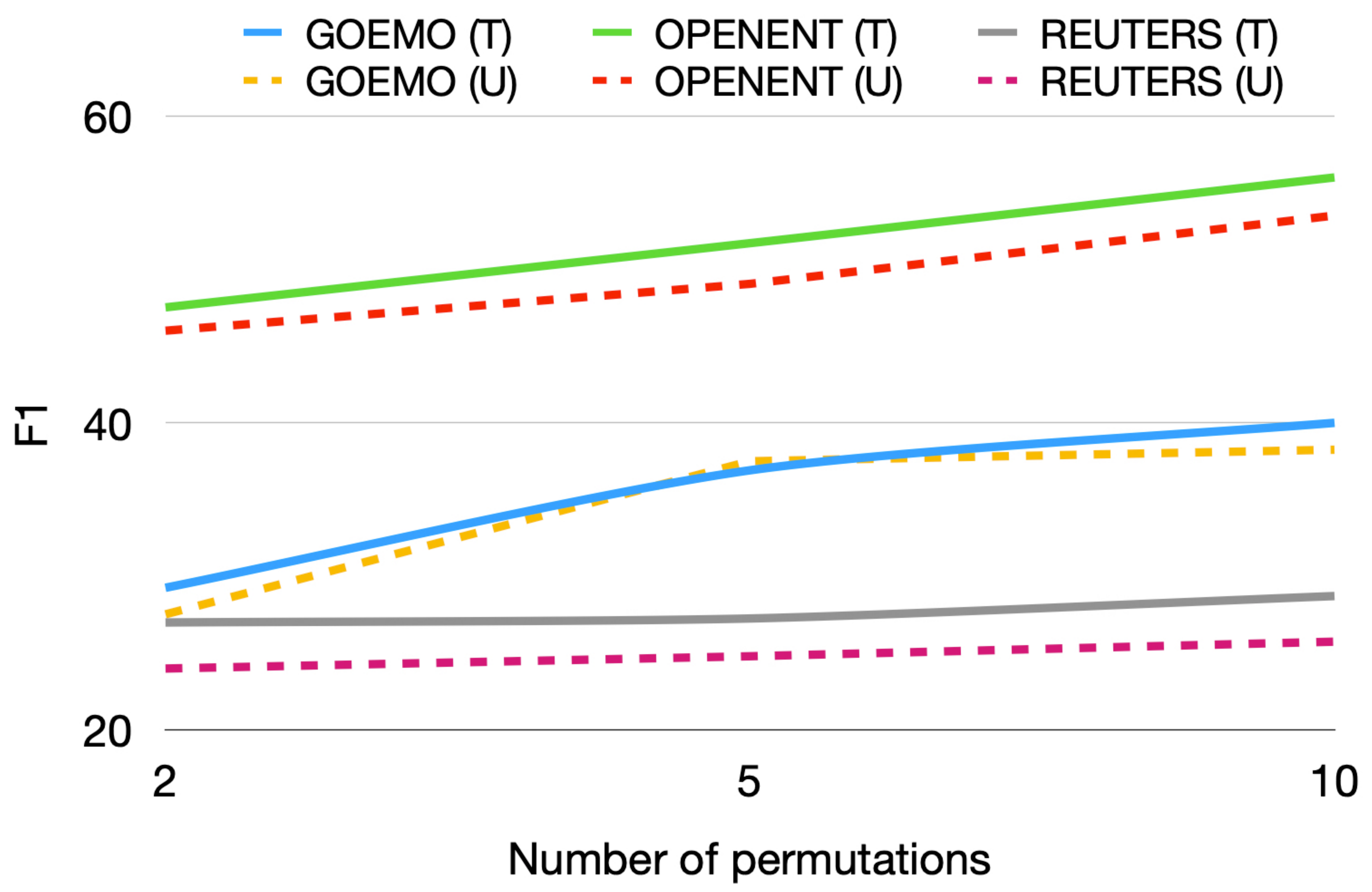}
\captionof{figure}{\ours (T) consistently outperforms \rand (U) as the number of permutations~($n$) is increased.}
\label{fig:tsampletreeRight}
\end{figure}



\subsubsection{Ablation study}
\label{sec:ablation}
We ablate the two critical components of our system: cardinality~(\ours\xspace - \card) and order~(\ours\xspace - \sample) and investigate the performance for each of these settings using \bart for multi-label classification.
Table~\ref{tab:ablations} presents the results.
Both the components individually help, but a larger drop is seen by removing cardinality.
We also train using \rand orders, instead of \sample. \rand does not improve over \sts consistently (both with and without \card), showing that merely augmenting with random permutations does not help.
Further, Appendix~\ref{sec:additionalresultsappendix} 
shows that cardinality is useful even with \rand order.
\begin{table}[!h]
\centering
\small
\begin{tabular}{@{}p{2cm}lll@{}}
\toprule & \emoclass   & \openent  & \reuters \\ \midrule
\ours & \textbf{30.0} & \textbf{53.5} & \textbf{26.7} \\
- \card  & 23.3 (-22\%) & 48.0 (-10\%) & 15.8 (-40\%)     \\ 
- \sample & 26.8 (-11\%) & 50.5 (-6\%) & 24.3 (-9\%)     \\

\rand & 27.5 (-8\%) & 50.4 (-6\%)  & 24.7 (-7\%)      \\ 
\textsc{freq} & 19.03 (-36\%) & 49.9 (-7\%)  & 23.4 (-12\%)      \\ 

\bottomrule
\end{tabular}%
\caption{Ablations: modeling cardinality~(\card) and 
sampling informative orders (\sample) both help,
with larger gains from \card. 
\rand ordering hurts.
}
\label{tab:ablations}
\end{table}

\subsubsection{Role of order}

\paragraph{Nature of permutations created by \ours}

\ours encourages highly co-occurring pairs ($\ry_i, \ry_j$) to be in the order $\ry_i, \ry_j$ if $\pcond{\ry_j}{\ry_i} > \pcond{\ry_i}{\ry_j}$.
In our analysis, this dependency in the datasets shows that the orders exhibit a pattern where \textit{specific} labels appear before the \textit{generic} ones.
E.g., in entity typing, the more generic entity \textit{event} is generated after the more specific entities \textit{home game} and \textit{match}~(see \Figref{fig:tsampletree}).

\paragraph{Increasing \# permutations ($n$) helps:} 
Fig. \ref{fig:tsampletreeRight} shows that \ours and \rand improve as $n$ is increased from $n=2$ to $10$; \ours outperforms \rand across $n$.

\paragraph{Reversing the order hurts performance} In order to check our hypothesis of whether only informative orders helping with set generation, we invert the label dependencies returned by \ours for all the datasets and train with the same model settings.
Across all datasets, we observe that reversing the order leads to an average of 12\% drop in \fs score. 
The reversed order not only closes the gap between \ours and \rand, but in many instances, the performance is slightly worse than \rand.

\paragraph{Ordering by frequency}
\emnlpcr{\citet{yang2018sgm} use frequency ordering, where the most frequent label is placed first in the sequence. We compare with this baseline in Table~\ref{tab:ablations}~(\textsc{freq}). The results indicate that the performance of frequency-based ordering is dataset dependent.
Relying on a fixed criteria like frequency might lead to skewed outputs, especially for  datasets with a long-tail of labels. 
For instance, for \openent, one of the most significant failure modes of the freq method was generating the most common label in the corpus (person) for every input. TSAMPLE can be seen as a way to balance the most frequent and least frequent labels in the corpus using \textsc{pmi} and conditional likelihood (Algorithm~\ref{alg:topographcreation}, L3).}

\subsubsection{Role of cardinality}
\paragraph{Cardinality is successfully predicted and used}
Table~\ref{tab:ablations} shows that cardinality is crucial to modeling set output. 
To study whether the models learn to condition on predicted 
cardinality,
we compute an \textit{agreement} score - defined as the \% of times the predicted cardinality matches the number of elements generated by the model.
The model effectively predicts the cardinality almost exactly in \emoclass and \reuters datasets~(avg. 95\%). 
While the exact match agreement is low in \openent~(35\%), the model is within an error of $\pm$1 in 93\% of the cases.
These results show that cardinality 
predicts the end of sequence (\textsc{eos}) token.
\emnlpadd{The accuracy for predicting the exact cardinality is 61\% across datasets, and it increases to 76\% within an error of 1 (\ie when the predicted cardinality is off by 1).}

\paragraph{Information about cardinality improves multi-label classification}
\clsbaseline baseline uses different values of k for predicting labels.
To test if knowledge of cardinality improves multi-class classification, we experiment with a setting where the true cardinality is available at inference~(\ie $k$ is set to the true value of cardinality).
Table~\ref{tab:oraclek} shows that 
cardinality improves performance.

\begin{table}[!h]
\centering
\small
\begin{tabular}{p{0.31\columnwidth}|p{0.17\columnwidth}|p{0.17\columnwidth}|p{0.17\columnwidth}}
\midrule
& \emoclass                 & \openent                  & \reuters \\ \midrule
\clsbaseline &  \textbf{22.4} & 14.3  & 21.7 \\
\clsbaselinewithk  & 21.3\tiny{(-4.9\%)} & \textbf{17.8}\tiny{(+24.5\%)} & \textbf{25.6}\tiny{(+18\%)}   \\\bottomrule
\end{tabular}%
\caption{Cardinality improves multi-label classification.}
\label{tab:oraclek}
\end{table}




\subsubsection{Error analysis}
\label{sec:erroranalysis}
We manually compare the outputs generated by the vanilla \bart model with \bart + \ours.
For the open-entity typing dataset, we randomly sample 100 examples and find that vanilla \sts approach generates sets with an ill-formed element 22\% of the time, whereas \ours completely avoids this. Examples of such ill-formed elements include \textit{personformer}, \textit{businessirm}, \textit{polit}, \textit{foundationirm}, \textit{politplomat}, \textit{eventlete}. This analysis indicates that training the model with an informative order infuses more information about the underlying type-hierarchy, avoiding the ill-formed elements.


\section{Related work}

\vspace{-0.5em}
\paragraph{Set generation in \nlp} 
Prior work has noted the impact of the order on the performance of text generation models~\citep{vinyals2015order}, especially in the context of keyphrase generation~\citep{meng2019doesordermatter}. Approaches to explicitly model set properties for NLP tasks include either performing an exhaustive search 
to find the best order~\citep{vinyals2015order}, changing the model training to modify the loss function~\citep{qin2019adapting}, or applying post-processing~\citep{chowdhury2016know2look}.
\emnlpadd{Notably, \citet{ye2021one2set} introduced One2Set, a method for training order-invariant models for generating set of keyphrases. Our main goal in this work is to provide a framework to identify useful orders for set generation, and show that such orders can help vanilla \sts models.}
\ours can work with any \sts model, and is complementary to these specialized methods.

\vspace{-0.5em}
\paragraph{Non-\sts set generation}
These include reinforcement learning for multi-label classification~\citep{yang2019deep} and combinatorial problems~\citep{nandwani2020neural}, and using pointer networks for keyphrase extraction~\citep{ye2021one2set}.
We focus on optimally adapting existing \sts models for set generation, without external knowledge~\citep{wang2020k,zhang2019ernie}.

\citet{chen2021order} explored the generation of an optimal order for graph generation \textit{given} the nodes.
They observed that ordering nodes before inducing edges improves graph generation.
However, in our case, since the labels are unknown, conditioning on the labels to create the optimal order is not possible.

\vspace{-0.5em}
\paragraph{Connection with Janossy pooling} \citet{murphy2019janossy} generalize deep sets by encoding a set of $N$ elements by pooling permutations of $P(N, k)$ tuples.
With $k=N$, their method is the same as pooling all $N!$ sequences, and reduces to deep sets with $k=1$. We share similarity with tractable searching over $N!$ with Janossy pooling, but instead of iterating over all possible 2-tuples, we impose pairwise constraints on the element order.

\vspace{-0.5em}

\paragraph{Modeling set input}
A number of techniques have been proposed for encoding set-shaped inputs~\citep{santoro2017simple,zaheer2017deep,lee2019set,murphy2019janossy,huangbetter2020,kim2021setvae}. 
Specifically, \citet{zaheer2017deep} propose deep sets, wherein they show that pooling the representations of individual set elements and feeding the resulting features to a non-linear network is a principled way of representing sets.
\citet{lee2019set} present permutation-invariant attention to encode shapes and images using a modified version of attention~\citep{vaswani2017attention}.
Unlike these works, we focus on settings where the input is a sequence, and the output is a set.


\section{Conclusion and Discussion}
We present \ours, a novel data augmentation method for conditional set generation that  incorporates informative orders and adds cardinality information. Our key idea is using the most likely order (vs. a randomly selected order) to represent a set as a sequence and conditioning the generation of a set on predicted cardinality.
As a computationally efficient and general-purpose plug-in data augmentation algorithm, \ours improves \sts models for set generation across a wide spectrum of tasks. For future work, it would be interesting to investigate if the general ideas in this work have applications in settings beyond set generation.
Examples include generating additional data to improve language modeling in low-resource scenarios and determining the ideal order of in-context examples in a prompt.

\section*{Acknowledgments}
We thank Amrith Setlur for thoughtful discussion and the anonymous reviewers for valuable feedback. 
This material is partly based on research sponsored in part by the Air Force Research Laboratory under agreement number FA8750-19-2-0200. 
The U.S. Government is authorized to reproduce and distribute reprints for Governmental purposes notwithstanding any copyright notation thereon. 
The views and conclusions contained herein are those of the authors and should not be interpreted as necessarily representing the official policies or endorsements, either expressed or implied, of the Air Force Research Laboratory or the U.S. Government.
We also thank Google for providing the TPUs for conducting experiments.
 
\newpage
\clearpage
\section*{Limitations}

\paragraph{Ineffectiveness on independent sets}  \ours is \textit{only} useful when the labels share some degree of dependence. For tasks where the labels are completely independent, \ours will not be effective. It can be shown that order will not affect learning joint distribution over labels if the labels are independent~(Lemma~\ref{sec:ordershouldnotmatter}). Thus, in such settings, \textit{any} method (including \ours) that seeks to leverage the relationship between labels will not be helpful. In addition to Lemma~\ref{sec:ordershouldnotmatter}, we conduct thorough simulation studies to verify this limitation~(Figure~\ref{fig:pplvsorderappendix}).

\paragraph{Use of large language models} We perform experiments with extremely large models, including T5-XXL and GPT-3 models. Particularly, GPT-3 is only available through OpenAI API; thus, all the details about its working are not publicly available. However, our experiments also show results using BART models that run on a single RTX 2080 GPU~(please also see details on reproducibility in Appendix~\ref{sec:reproducibility}).
\emnlpcr{Further, such language models are typically trained on a large English corpora, which is also the focus of our work.}

\paragraph{Focus on \sts} A key limitation of our work is that it focuses on set-generation using \sts models. Thus the proposed insights may not apply to other settings~(\eg computer vision) where using language models is not directly feasible.
Nevertheless, with the growing popularity of libraries like Huggingface~\citep{wolf2019huggingface}, we anticipate that \sts models will be applied to a growing number of use cases, even those that would traditionally be tackled using a non-\sts method.
Further, we compare our method with representative non-\sts baselines~(like multi-label classifier).

To our knowledge, our work does not directly use any datasets that contain explicit societal biases. Therefore, we anticipate that our work will not lead to any significant negative implications concerning real-world applications.

\bibliography{main}
\bibliographystyle{acl_natbib}
\newpage
\clearpage
\appendix
\label{sec:appendix}

\section{Reproducibility}
\label{sec:reproducibility}
We take the following steps for reproducibility of our results:
\begin{enumerate}
    \item All the experiments are performed for three different random seeds. In addition, we conduct a proportion of samples hypothesis test to establish the statistical significance of our results. We did not perform extensive hyperparameter tuning and used the same set of defaults for baselines and our proposed method.
    \item For all data augmentation experiments, we match the number of training samples and epochs; all the models are trained for the same duration. This alleviates the concern that the models perform well with augmented data merely because of the longer training time.
    \item We conduct a proportion of samples test for all the experiments conducted on real-world datasets and use a small $p=0.0005$ to measure highly significant results, which are indicated with an underscore.
\end{enumerate}

Our work aims to promote the usage of existing resources for as many use cases as possible. In particular, all our experiments are performed on the BASE-version of the model (BART) that can relatively lower parameter count to conserve resources and help lower our impact on climate change.

\section{Proofs}
Let $\sY$ be the output space, $\ry_i, \ry_j, \ry_k \in \sY$, and $\vy_k \in \sY - \ry_i - \ry_j$ be a subset of the symbols excluding $\ry_i, \ry_j$. We assume that all the distributions are non-negative~(\ie $p(\vy) > 0, \forall \vy \in \sY$)
\begin{lemma}
$\ry_i \not \ind \ry_j \implies \ry_i \not \ind (\ry_j \ry_k)$
\label{sec:higherorderdeps}
\end{lemma}
\paragraph{Proof} Let $\ry_i \ind (\ry_j \ry_k)$ by contradiction. Then:
\begin{align}
    p(\ry_i, \ry_j \ry_k) &= p(\ry_i)p(\ry_j \ry_k) 
    \label{eqn:contradiction}
\end{align}
Also, 
\begin{align}
    p(\ry_i, \ry_j) &= \sum_{\ry_k \in \mZ} p(\ry_i, \ry_j \ry_k) \nonumber \\
    &= \sum_{\ry_k \in \mZ} p(\ry_i)p(\ry_j \ry_k) \nonumber \tag{\eqref{eqn:contradiction}} \\
    &=  p(\ry_i)\sum_{\ry_k \in \mZ}p(\ry_j \ry_k) \nonumber \\
    &= p(\ry_i)p(\ry_j)
\end{align}
However, $\ry_i \not \ind \ry$ thus $\ry_i \not \ind \ry \implies \ry_i \not \ind (\ry_j \ry_k)$.





\begin{lemma}
$$ \pcond{\ry_i}{\ry_j} > \pcond{\ry_j}{\ry_i} \implies \pcond{\ry_i}{\ry_j, \vy_k} > \pcond{\ry_j}{\ry_i, \vy_k}$$
\text{if} $\ry_i, \ry_j \ind \vy_k$
\label{sec:conditioningretainsprobs}
\end{lemma}

\paragraph{Proof} We have:
\begin{align}
  \pcond{\ry_i}{\ry_j} &> \pcond{\ry_j}{\ry_i} \nonumber \\
  \implies p(\ry_j) &< p(\ry_i)
  \label{eqn:p5lessthan}
\end{align}

\begin{align}
    p(\ry_j, \vy_k) &= \pcond{\vy_k}{\ry_j}p(\ry_j) \nonumber \\
    &< \pcond{\vy_k}{\ry_j}p(\ry_i) \tag{Equation~\ref{eqn:p5lessthan}} \nonumber \\
   &= \pcond{\vy_k}{\ry_i}p(\ry_i) \nonumber \tag{$\ry_i, \ry_j \ind \vy_k \implies \pcond{\vy_k}{\ry_j} = \pcond{\vy_k}{\ry_i} = p(\vy_k)$} \\
   &= p(\ry_i, \vy_k) 
   \label{eqn:p5jklessik}
\end{align}

Thus, 
\begin{align}
    \pcond{\ry_i}{\ry_j, \vy_k} &= \frac{p(\ry_i, \ry_j, \vy_k)}{p(\ry_j, \vy_k)} \nonumber \\ &> \frac{p(\ry_i, \ry_j, \vy_k)}{p(\ry_i, \vy_k)} \nonumber \\
    &= \pcond{\ry_j}{\ry_i, \vy_k}
    \label{eqn:conditioning}
\end{align}

\begin{lemma}
If $\ry_i \ind \ry_j\ \forall \ry_i, \ry_j \in \sY$, the order is guaranteed to not affect learning.
\label{sec:ordershouldnotmatter}
\end{lemma}
\paragraph{Proof} 
Let $\pi_j$ be the $j^{th}$ order over $\sY$ (out of $|\sY|!$ possible orders $\Pi$), and $\pi_j(\sY)$ be the sequence of elements in $\sY$ arranged with $\pi_j$.
\begin{align*}
  \pcond{\ry_i}{\ry_j} &= p(\ry_i) \tag{$\ry_i \ind \ry_j\ \forall \ry_i, \ry_j$}  \\
    \implies p(\ry_i, \ry_j, \ry_k) &= p(\ry_i)\pcond{\ry_j}{\ry_i}\pcond{\ry_k}{\ry_i,\ry_j}\\
    &= p(\ry_i)p(\ry_j)p(\ry_k) \\
\implies p(\pi_m(\ry_i, \ry_j, \ry_k)) &= p(\pi_n(\ry_i, \ry_j, \ry_k))\ \forall \pi_m, \pi_m \in \Pi
\end{align*}
In other words, when all elements are mutually independent, all possible joint factorizations will simply be a product of the marginals, and thus identical.

\begin{lemma}
The graphs constructed to sample orders for \ours cannot have cycles. 
\label{sec:nocycles}
\end{lemma}
\paragraph{Proof}
Let $\ry_i, \ry_j, \ry_k$ form a cycle: $\ry_i \rightarrow \ry_j  \rightarrow \ry_k \rightarrow \ry_i$.
By construction, the following conditions must hold for such a cycle to be present:
\begin{align*}
    \log \pcond{\ry_j}{\ry_i} - \log \pcond{\ry_i}{\ry_j} &> \beta \implies \log p(\ry_i) < \log p(\ry_j) \\
    \log \pcond{\ry_k}{\ry_j} - \log \pcond{\ry_j}{\ry_k} &> \beta \implies \log p(\ry_j) < \log p(\ry_k) \\
    \log \pcond{\ry_i}{\ry_k} - \log \pcond{\ry_k}{\ry_i} &> \beta \implies \log p(\ry_k) < \log p(\ry_i) \\
\end{align*}
Putting the three implications together, we get $\log p(\ry_i) < \log p(\ry_j) < \log p(\ry_k)  < \log p(\ry_i)$, which is a contradiction. Hence, the graphs constructed for \ours cannot have a cycle.

\section{Sample graphs}
\label{sec:moreexamples}
\begin{figure}[!ht]
    \centering
  \includegraphics[width=\linewidth]{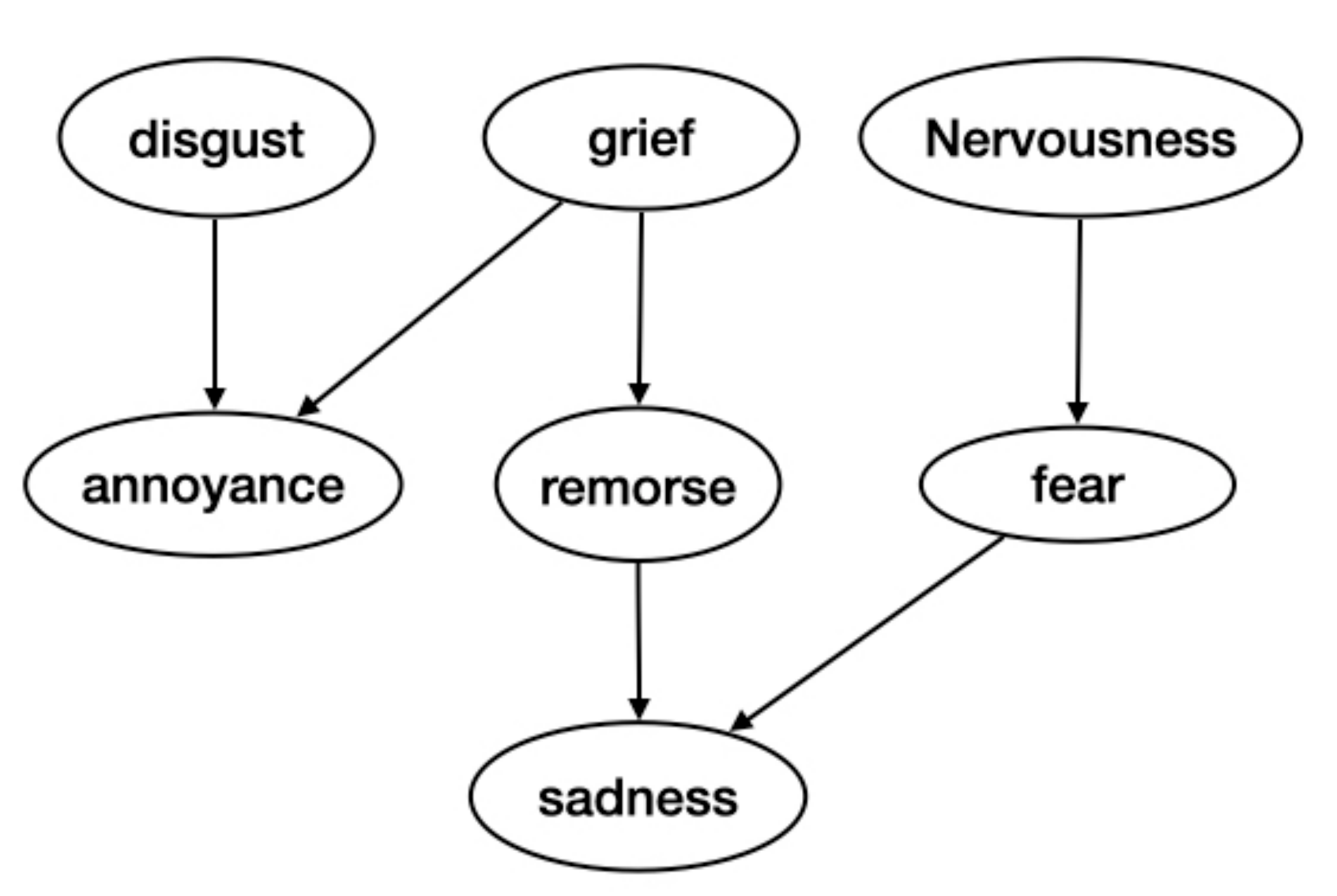}
    \caption{Label dependencies discovered by \sample for \emoclass}
    \label{fig:tsamplemoreexamplesemo}
\end{figure}

\begin{figure}[!ht]
    \centering
  \includegraphics[width=0.9\linewidth]{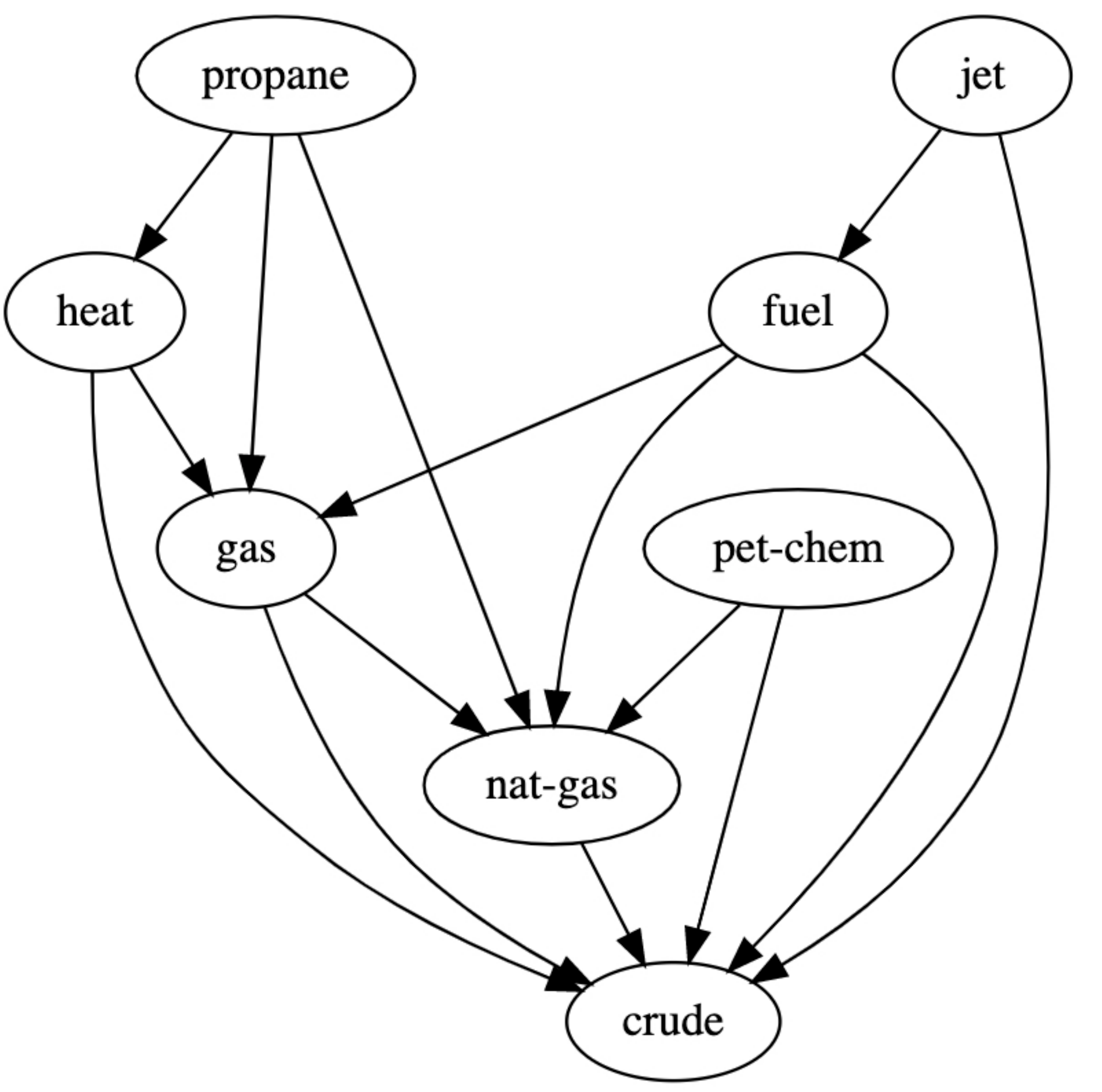}
    \caption{Label dependencies discovered by \sample for \reuters}
    \label{fig:tsamplemoreexamplesreuters}
\end{figure}

In this section, we present additional examples from \reuters and \emoclass datasets to illustrate the permutations generated by our method.
As discussed in~\Secref{sec:tsample}, \ours encourages highly co-occuring pairs ($\ry_i, \ry_j$) to be in the order $\ry_i, \ry_j$ if $\pcond{\ry_j}{\ry_i} > \pcond{\ry_i}{\ry_j}$.
In our analysis, this dependency in the datasets shows that the orders exhibit a pattern where \textit{specific} labels appear before the \textit{generic} ones.
For example, in case of entity typing, the more \emoclass, \textit{sadness} is generated after the more specific emotion \textit{remorse} and \textit{fear}~(\Figref{fig:tsamplemoreexamplesemo}).
Similarly, the entity \textit{crude} is generated after the entities \textit{gas} and \textit{nat-gas}.~(\Figref{fig:tsamplemoreexamplesreuters} (right)).

\section{Hyperparameters}
\label{sec:hyperparams}

We list all the hyperparameters in Table~\ref{tab:hyperparams}.

\begin{table}[!ht]
\small
\centering
\begin{tabular}{@{}lc@{}}
\toprule
Hyperparameter            & Value                      \\ \midrule
GPU                       & GeForce RTX 2080 Ti        \\
gpus                      & 1                          \\
auto\_select\_gpus        & false                      \\
accumulate\_grad\_batches & 1                          \\
max\_epochs               & 3                          \\
precision                 & 32                         \\
learning\_rate            & 1e-05                      \\
adam\_epsilon             & 1e-08                      \\
num\_workers              & 16                         \\
warmup\_prop              & 0.1                        \\
seeds                     & {[}15143, 27122, 999888{]} \\
add\_lr\_scheduler        & true                       \\
lr\_scheduler             & linear                     \\
max\_source\_length       & 120                        \\
max\_target\_length       & 120                        \\
val\_max\_target\_length  & 120                        \\
test\_max\_target\_length & 120 \\ \bottomrule
\end{tabular}
\caption{List of hyperparameters used for all the experiments.}
\label{tab:hyperparams}
\end{table}

\section{Dataset}
\label{dataset:examples}

Table~\ref{tab:datasetexamples} shows examples for each of the datasets.

\begin{table*}[ht!]
\centering
\begin{tabular}{@{}lll@{}}
\toprule
 & Input & Output \\ \midrule
\begin{tabular}[c]{@{}l@{}}Fine-grained emotion\\ classification, [28]\\~\citep{demszky2020goemotions} \end{tabular} &
  \begin{tabular}[c]{@{}l@{}}\textit{So there's hope for the rest of us!} \\ \textit{Thanks for sharing. What helped}\\ \textit{you get to where you are?}\end{tabular} &
  \begin{tabular}[c]{@{}l@{}}\{curiosity, gratitude, \\ optimism\}\end{tabular} \\ \midrule
Open-entity typing [2519]\\~\citep{choi2018ultra} &
  \begin{tabular}[c]{@{}l@{}}\textit{Some 700,000 cubic meters of} \\ \textit{caustic sludge and water burst} \\ \textit{inundating $[\text{SPAN}]$ three west}\\
  \textit{Hungarian villages $[\text{SPAN}]$ and spilling.}\end{tabular} &
  \begin{tabular}[c]{@{}l@{}}\{colony, region,\\ location,  hamlet,\\ area, village,\\ settlement, community\}\end{tabular} \\ 
  \midrule
Reuters [90]\\~\citep{lewis1997reuters} &
  \begin{tabular}[c]{@{}l@{}}\textit{India is reported to have bought} \\ \textit{two white sugar cargoes for$\ldots$} \\ \textit{$\ldots$cargo sale, they said.}\end{tabular} &
  \begin{tabular}[c]{@{}l@{}}\{ship, sugar\}\end{tabular} \\  \midrule 
 Keyphrase generation [270k]~\\\citep{ye2021one2set} &
  \begin{tabular}[c]{@{}l@{}}We analyze the impact of core \\ affinity on both network and \\ disk i/o performance...our dynamic\\ core affinity improves the file upload\\ throughput more than digit\%\end{tabular} &
  \begin{tabular}[c]{@{}l@{}}\{big data, multi-core,\\  process-scheduling\}\end{tabular} \\
  \bottomrule
\end{tabular}
\caption{Real world tasks used for experiments}
\label{tab:datasetexamples}
\end{table*}

\newpage
\clearpage

\section{Additional results}
\label{sec:additionalresultsappendix}
This section presents detailed results that were omitted from the main paper for brevity.
This includes macro and micro precision, recall, and \fst scores on all datasets, and additional ablation experiments.
\begin{enumerate}
    \item Table~\ref{tab:mainresultsdetailed} shows the detailed results from the four tasks.
    \item Detailed results on \emoclass, \reuters, and \openent are present in Tables~\ref{tab:allresultsgo}, \ref{tab:allresultsreuters}, and \ref{tab:allresultsopenent}, respectively.
    \item Table~\ref{tab:resultsv3withclassifieradditional} includes results from a multi-label classification baseline where bert-base-uncased is used as the encoder.
\end{enumerate}

\begin{table}[]
\small
\centering
\addtolength{\tabcolsep}{-3pt}    
\begin{tabular}{@{}lrrr@{}}\toprule
                           & \emoclass     & \openent      & \reuters      \\ \midrule
\clsbaseline & 22.4          & 14.3          & 21.7          \\
\clsbaseline@oracle-k  & 21.3          & 17.8          & 25.6          \\
\ours + card            & \textbf{30.0} & \textbf{53.5} & \textbf{26.7} \\ \bottomrule
\end{tabular}%
\addtolength{\tabcolsep}{3pt}
\caption{Multi-label classification when the true cardinality is provided to the classifier. While providing the true cardinality helps the performance of multi-label classifiers, it still lags \ours.}
\label{tab:oraclekresultsdetailed}
\end{table}

\setlength{\tabcolsep}{.7em}
\begin{table*}[ht]
\small
\centering
\begin{tabular}{@{}lrrrrrrrrcrrrrrr@{}}
\toprule
          & \multicolumn{3}{c}{\emoclass} & \multicolumn{3}{c}{\openent} & \multicolumn{3}{c}{\reuters} & \multicolumn{3}{c}{\keygen} \\ \midrule
 &
  \multicolumn{1}{c}{\pre} &
  \multicolumn{1}{c}{\rec} &
  \multicolumn{1}{c}{\fst} &
  \multicolumn{1}{c}{\pre} &
  \multicolumn{1}{c}{\rec} &
  \multicolumn{1}{c}{\fst} &
  \multicolumn{1}{c}{\pre} &
  \multicolumn{1}{c}{\rec} &
  \multicolumn{1}{c}{\fst} &
    \multicolumn{1}{c}{\pre} &
  \multicolumn{1}{c}{\rec} &
  \multicolumn{1}{c}{\fst} & \\ \midrule
\clsbaseline  & 20.8 & 42.4  & 22.4 & 16.4 & 25.1 & 14.3 & 19.7 & 43.4 & 21.7 & - & - & -   \\
\clsbaselinewithk  & 21.3&	21.3	&21.3	&17.8&	17.8&	17.8&	25.6&	25.6&	25.6& - & - & -     \\
\midrule
\bsearch  & 10.7 &	7.0 & 7.4    & 26.5     & 31.4    & 26.3    & 10.9     & 7.1     & 7.5 & 5.8	& 	\textbf{7.4} & 6.4    \\
\baseline & 27.4 &	26.2	& 23.4    & 55.4     & 42.4    & 44.6    & 24.8 &	13.8 &	15.6 & 6.7 &		5.5	 &	5.9   \\
\rand  & 32.5 &	19.9 &	22.7   & 62.6     & 41.7    & 46.9    &  26.7     & 12.7    & 15.2  & 6.6 &		4.5 &		5.2    \\
\ours  & \textbf{36.7} &	19.8	& 23.3   & 60.0    & 44.5    & 48.0   &  26.5	& 12.8	& 15.8  & 7.0	 & 	5.0	 & 	5.6   \\
\midrule
\sts + \card     & 33.0	& 28.3 &	26.8    & 62.5     & 44.7    & 50.5    & 34.1	& 21.8 &	24.3 & 7.1	 & 	5.6	 & 	6.1      \\
\rand + \card   & 35.6&	26.5&	27.5    & \textbf{\emph{68.6}}     & 42.3    & 50.4    & 35.3	& 22.1	& 24.7 & 7.3 & 		5.7 & 		6.3     \\
\ours + \card &  36.1 &	\textbf{30.5} &	\textbf{30.0}    & \emph{65.5}     & \textbf{\emph{47.5}}    & \textbf{\emph{53.5}}    & \textbf{\emph{36.7}}	& \textbf{\emph{24.1}} &	\textbf{\emph{26.7}} & \textbf{7.7}	&	6.1 &	\textbf{6.6}     \\ \bottomrule
\end{tabular}
\caption{Our main results in detail: using permutations generated by \ours and adding cardinality gives the best overall performance in terms of macro precision, recall, and \fs score. \clsbaseline     is the standard multi-label classification approach. Statistically significant results~(\textbf{p = }$5e^{-4}$) are \emph{underscored}. \card stands for cardinality.}
\label{tab:mainresultsdetailed}
\end{table*}

\setlength{\tabcolsep}{0.5em}

\begin{table*}[ht]
\centering
\begin{tabular}{@{}lrrrrrrr@{}}
\toprule
 &
  \multicolumn{1}{l}{\pmicro} &
  \multicolumn{1}{l}{\pmacro} &
  \multicolumn{1}{l}{\rmicro} &
  \multicolumn{1}{l}{\rmacro} &
  \multicolumn{1}{l}{\fmicro} &
  \multicolumn{1}{l}{\fmacro} &
  \multicolumn{1}{l}{\jaccard} \\ \midrule
\bsearch               & 47.17 & 10.68 & 13.09 & 7.02  & 10.7 & 7.36  & 7.4  \\
\sts                   & 41.65 & 27.39 & 35.19 & 26.21 & 27.4 & 23.41 & 23.4 \\
\sts + \card              & 39.77 & 33    & 38.02 & 28.31 & 33   & 26.79 & 26.8 \\
\rand + \card             & 44.77 & 35.6  & 32.96 & 26.54 & 35.6 & 27.53 & 27.5 \\
\ours + \card & 43.37 & 36.08 & 34.51 & 30.54 & 36.1 & 30.01 & 30   \\
\rand - \card           & 48.85 & 32.45 & 27.75 & 19.86 & 32.5 & 22.67 & 22.7 \\
\ours - \card & 50    & 36.68 & 29.84 & 19.84 & 36.7 & 23.31 & 23.3 \\ \bottomrule
\end{tabular}
\caption{Results for \emoclass.}
\label{tab:allresultsgo}
\vspace{2em}
\begin{tabular}{@{}lrrrrrrr@{}}
\toprule
\textbf{} &
  \multicolumn{1}{l}{\pmicro} &
  \multicolumn{1}{l}{\pmacro} &
  \multicolumn{1}{l}{\rmicro} &
  \multicolumn{1}{l}{\rmacro} &
  \multicolumn{1}{l}{\fmicro} &
  \multicolumn{1}{l}{\fmacro} &
  \multicolumn{1}{l}{\jaccard} \\ \midrule
\bsearch               & 70.04 & 10.92 & 34.9  & 7.1   & 46.56 & 7.54  & 37.49 \\
\sts                   & 66.36 & 24.74 & 42.28 & 13.78 & 51.64 & 15.58 & 44.3  \\
\sts + \card              & 73.02 & 34.17 & 53.8  & 21.85 & 61.95 & 24.28 & 59.08 \\
\rand + \card             & 74.26 & 35.31 & 54.33 & 22.13 & 62.75 & 24.74 & 58.95 \\
\ours + \card  & 75.65 & 36.67 & 55.54 & 24.13 & 64.05 & 26.66 & 61.14 \\
\rand - \card           & 69.56 & 26.68 & 38.15 & 12.71 & 49.27 & 15.2  & 42.24 \\
\ours - \card & 76.55 & 26.49 & 41.78 & 12.77 & 54.06 & 15.78 & 47.34 \\ \bottomrule
\end{tabular}
\caption{Results for \reuters.}
\label{tab:allresultsreuters}
\vspace{2em}
\centering
\begin{tabular}{@{}lrrrrrrr@{}}
\toprule
\textbf{} &
  \multicolumn{1}{l}{\pmicro} &
  \multicolumn{1}{l}{\pmacro} &
  \multicolumn{1}{l}{\rmicro} &
  \multicolumn{1}{l}{\rmacro} &
  \multicolumn{1}{l}{\fmicro} &
  \multicolumn{1}{l}{\fmacro} &
  \multicolumn{1}{l}{\jaccard} \\ \midrule
\bsearch               & 24.65 & 26.5  & 29.98 & 31.44 & 23.92 & 26.25 & 13.39 \\
\sts                   & 52.78 & 55.4  & 39.84 & 42.42 & 41.45 & 44.63 & 24.6  \\
\sts + \card              & 61.26 & 62.48 & 41.87 & 44.68 & 48.07 & 50.48 & 27.84 \\
\rand + \card             & 67.56 & 68.59 & 39.61 & 42.25 & 47.98 & 50.4  & 26.89 \\
\ours + \card     & 64.58 & 65.53 & 44.6  & 47.46 & 51.2  & 53.48 & 29.39 \\
\rand - \card           & 60.93 & 62.57 & 39.09 & 41.69 & 44.2  & 46.85 & 25.26 \\
\ours - \card      & 58.02 & 59.88 & 42.63 & 44.95 & 46.54 & 48.86 & 26.82 \\
\bottomrule
\end{tabular}
\caption{Results for \openent.}
\label{tab:allresultsopenent}
\end{table*}


\setlength{\tabcolsep}{.7em}
\begin{table*}[ht]
\centering
\begin{tabular}{@{}lrrrrrrrrcrrr@{}}
\toprule
          & \multicolumn{3}{c}{\emoclass} & \multicolumn{3}{c}{\openent} & \multicolumn{3}{c}{\reuters} \\ \midrule
 &
  \multicolumn{1}{c}{\pre} &
  \multicolumn{1}{c}{\rec} &
  \multicolumn{1}{c}{\fst} &
  \multicolumn{1}{c}{\pre} &
  \multicolumn{1}{c}{\rec} &
  \multicolumn{1}{c}{\fst} &
  \multicolumn{1}{c}{\pre} &
  \multicolumn{1}{c}{\rec} &
  \multicolumn{1}{c}{\fst} \\ \midrule
\bert @1  & 31.8 & 10.3 & 15.6  & 38.0 & 10.3 & 15.9    & 31.7 & 12.3 & 17.6     \\
\bert@3  & 23.8 & 23.4 & 23.6   & 19.7 & 14.0 & 16.1    & 23.4 & 28.3 & 25.5     \\
\bert@5  & 20.6 & 34.0 & 25.7   & 15.5 & 18.0 & 16.4    & 18.8 & 37.6 & 24.9     \\
\bert@10  & 16.5 & 54.3 & 25.3   & 11.8 & 26.0 & 16.0    & 15.1 & 61.8 & 24.2     \\
\bert@20  & 14.1 & 93.2 & 24.5    & 8.4 & 34.3 & 13.5    & 9.5 & 75.9 & 16.8     \\
\bert@50  & - &	- & -    & 2.6 & \textbf{50.2} & 4.9    &   8.9  & - &	- & -    \\
\bert  & 21.4 & 43.0 & 22.9 & 16.0 & 25.5 & 13.8 & 19.7 & 43.2 & 21.8    \\
\midrule
\bart@1  & 31.7 &	10.3 & 15.5   & 38.0    & 10.3    & 15.6    & 31.8     & 12.3     & 17.6     \\
\bart@3  & 21.2 &	21.0 & 21.0    & 19.7     & 14.0    & 15.8    & 23.1     & 28.1     & 25.2     \\
\bart@5  & 14.1 &	33.4 & 25.6    & 15.5     & 18.0    & 16.2    & 18.7     & 37.6     & 24.8     \\
\bart@10  & 16.3 &	53.4 & 25.0    & 11.7     & 26.0    & 15.9    & 15.1     & 62.0     & 24.1     \\
\bart@20  & 14.1 &	\textbf{93.3} & 24.5    & 8.4     & 34.3    & 13.4    & 9.6     & \textbf{77.1}    & 17.1     \\
\bart@50  & - &	- & -    & 4.9     &  48.0     &   8.9  & - &	- & -    \\
\bart  & 20.8 & 42.4  & 22.4 & 16.4 & 25.1 & 14.3 & 19.7 & 43.4 & 21.7    \\
\midrule
\bsearch  & 10.7 &	7.0 & 7.4    & 26.5     & 31.4    & 26.3    & 10.9     & 7.1     & 7.5     \\
\baseline & 27.4 &	26.2	& 23.4    & 55.4     & 42.4    & 44.6    & 24.8 &	13.8 &	15.6    \\
\rand  & 32.5 &	19.9 &	22.7   & 62.6     & 41.7    & 46.9    &  26.7     & 12.7    & 15.2      \\
\ours  & \textbf{36.7} &	19.8	& 23.3   & 60.0    & 44.5    & 48.0   &  26.5	& 12.8	& 15.8     \\
\midrule
\sts+\card     & 33.0	& 28.3 &	26.8    & 62.5     & 44.7    & 50.5    & 34.1	& 21.8 &	24.3      \\
\rand + \card   & 35.6&	26.5&	27.5    & \textbf{\emph{68.6}}     & 42.3    & 50.4    & 35.3	& 22.1	& 24.7      \\
\ours + \card &  36.1 &	\textbf{30.5} &	\textbf{30.0}    & \emph{65.5}     & \textbf{\emph{47.5}}    & \textbf{\emph{53.5}}    & \textbf{\emph{36.7}}	& \textbf{\emph{24.1}} &	\textbf{\emph{26.7}}      \\ \bottomrule
\end{tabular}
\caption{Our main results: using permutations generated by \ours and adding cardinality gives the best overall performance in terms of macro precision, recall, and \fs-score score. Statistically significant results are \emph{underscored}. \card stands for cardinality. \bert@k / \bart@k denotes the pointwise classification baseline using \bert / \bart where the top $k$ labels are used as the model output. The average is denoted by \bert / \bart.}
\label{tab:resultsv3withclassifieradditional}
\end{table*}

\newpage
\clearpage
\section{Exploring the influence of order on \sts models with a simulation}
\label{sec:simulationcomplete}
We design a simulation to investigate the effects of output order and cardinality on conditional set generation, following prior work that has found simulation to be an effective for studying properties of deep neural networks~\citep{vinyals2015order,khandelwal2018sharp}.

We note that a number of techniques have been proposed for encoding set-shaped inputs~\citep{santoro2017simple,zaheer2017deep,lee2019set,murphy2019janossy,huangbetter2020,kim2021setvae}.
Unlike these works, we focus on settings where the input is a sequence, and the output is a set, and design the data generation process accordingly.
%

\paragraph{Data generation}
We use a graphical model~(\Figref{fig:gmodelappendix}) to generate conditionally dependent pairs $(\vx, \sY)$, with different levels of interdependencies among the labels in $\sY$.
Let $\sY = \{\ry_1, \ry_2, \ldots, \ry_N\}$ be the set of output labels.
We sample a dataset of the form $\{(\vx, \vy)\}_{i=1}^{m}$.
$\vx$ is an $N$ dimensional multinomial sampled from a dirichlet parameterized by $\alpha$, and $\vy$ is a sequence of symbols with each $\evy_i \in \sY$.
The output sequence $\vy$ is created in $B$ \textit{blocks}, each block of size $k$.
A block is created by first sampling $k-1$ prefix symbols independently from $\text{Multinomial}(\vx)$, denoted by $\vy_{p}$
The $k^{th}$ suffix symbol ($y_{s}$) is sampled from either a uniform distribution with a probability = $\epsilon$ or is deterministically determined from the preceding $k-1$ prefix terms.
For block size of 1 ($k=1$), the output is simply a set of size $B$ sampled from $\vx$ (i.e., all the elements are independent). 
Similarly, $k=2$ simulates a situation with a high degree of dependence: each block is of size 2, with the prefix sampled independently from the input, and the suffix determined deterministically from the prefix.
Gradually increasing the block size increases the number of independent elements.

\begin{figure}[H]
  \centering
  \tikz{ %
    \node[obs] (alpha) {Dir($\alpha$)} ; %
    \node[latent, right=of alpha] (x) {$X$} ; %
    \node[latent, right=of x] (ypre) {$y_{p}$} ; %
    \node[latent, right=of ypre] (ysuff) {$y_{s}$} ; %
    \plate[inner sep=0.2cm, xshift=-0.12cm, yshift=0.12cm] {plate1} {(ypre)} {k-1}; %
    \plate[inner sep=0.2cm, xshift=-0.12cm, yshift=0.12cm] {plate2} {(ypre) (ysuff) (plate1)} {B}; %
    \plate[inner sep=0.2cm, xshift=-0.12cm, yshift=0.12cm] {plate3} {(x) (plate1) (plate2)} {M}; %
    \edge {alpha} {x} ; %
    \edge {x} {ypre} ; %
    \edge {ypre} {ysuff} ; %
  }
 \caption{The generative process for simulation}
 \label{fig:gmodelappendix}
\end{figure}
\begin{figure*}[ht]
    \centering
    \includegraphics[scale=0.4]{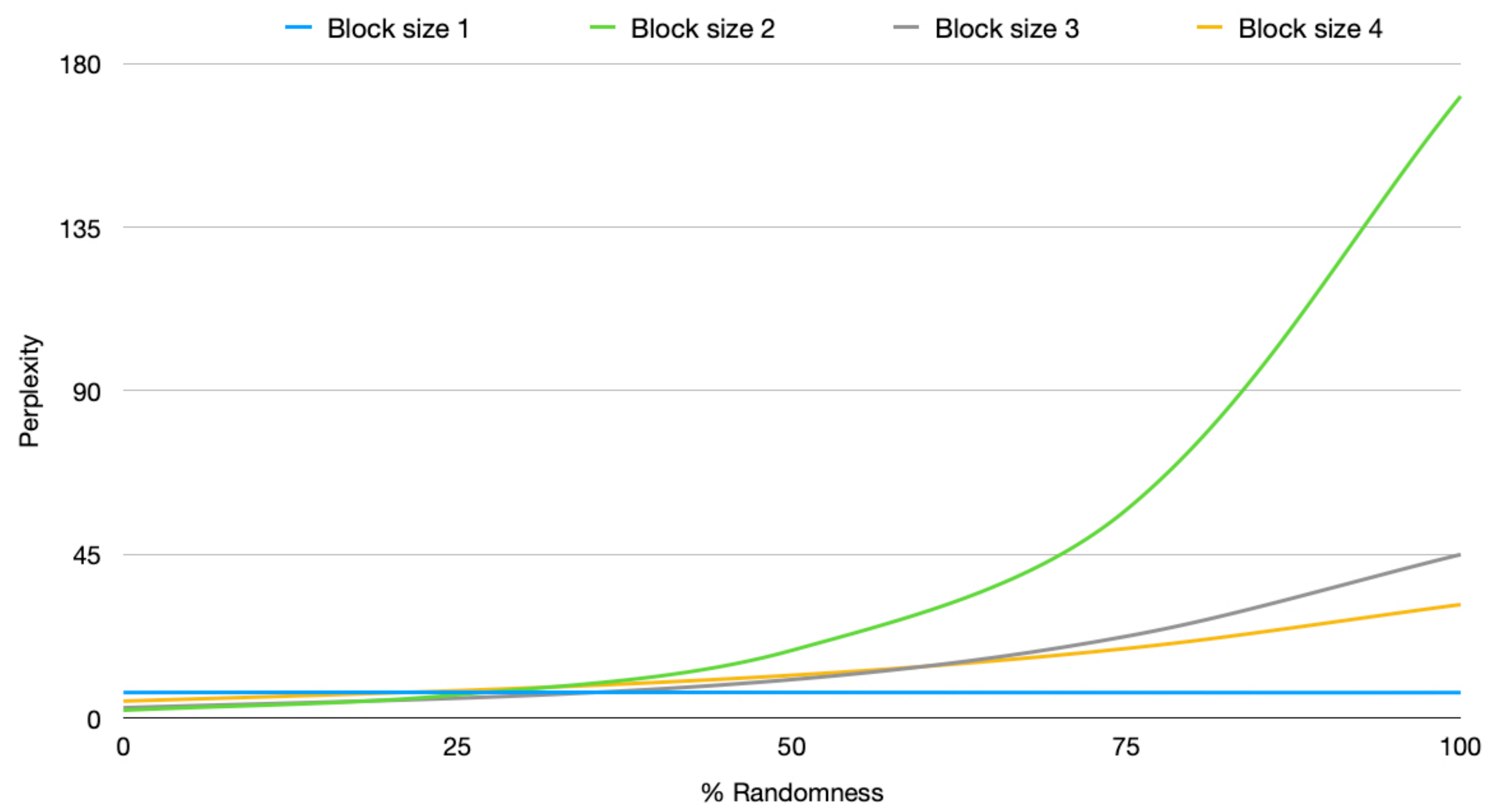}
    \caption{Perplexity vs. Randomness for varying block sizes. The degree of dependence between the labels is highest for block size = 2, where each label depends the preceding label. In such cases, the model is most affected by shuffling the order at test time. In contrast, with block size of 1, the perplexity is nearly unaffected by the order. This result complements Lemma~\ref{sec:ordershouldnotmatter} in showing that order will not affect \sts models if all the labels are independent of each other.}
    \label{fig:pplvsorderappendix}
\end{figure*}


\subsection{Major Findings}

We now outline our findings from the simulation.
We use the architecture of \bart-base~\cite{lewis2020bart}~(six-layers of encoder and decoder) without pre-training for all simulations.
All the simulations were repeated using three different random seeds, and we report the averages.
\paragraph{Finding 1: \sts models are sensitive to order, but only if the labels are conditionally dependent on each other.}
We train with the prefix $\vy_p$ listed in the lexicographic order. 
At test time, the order of is randomized from 0\% (same order as training) to 100 (appendixly shuffled).
As can be seen from~\Figref{fig:pplvsorderappendix} the perplexity gradually increases with the degree of randomness.
Further, note that perplexity is an artifact of the model and is independent of the sampling strategy used, showing that order affects learning.

\paragraph{Finding 2: Training with random orders makes the model less sensitive to order}

\begin{figure*}[!htb]
  \includegraphics[width=\linewidth]{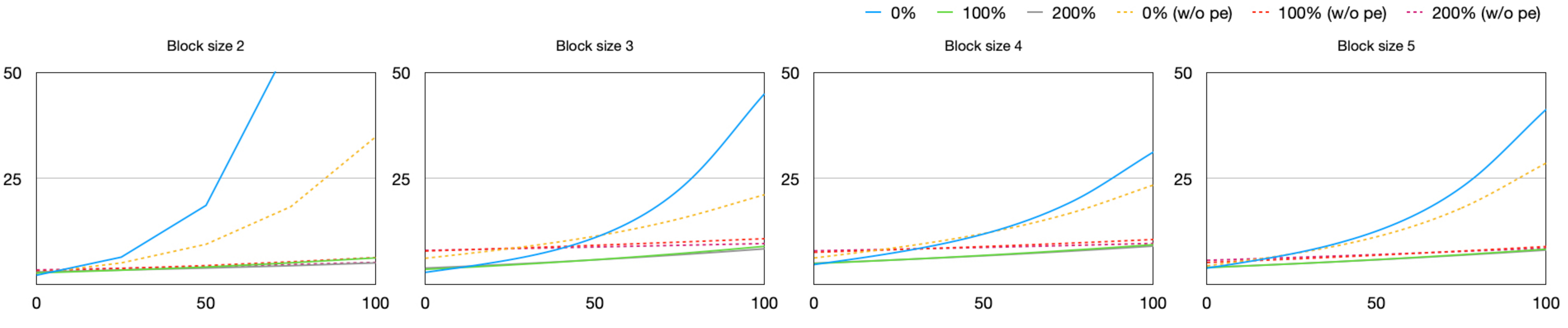}
\caption{Augmenting dataset with multiple orders help across block sizes. Augmentations also overcome any benefit that is obtained by using position embeddings.}
\label{fig:pplorderhelpsappendix}
\end{figure*}

As \Figref{fig:pplorderhelpsappendix} shows, augmenting with random order makes the model less sensitive to order.
Further, augmenting with random order keeps helping as the perplexity gradually falls, and the drop shows no signs of flattening.


\paragraph{Finding 3: Effects of position embeddings can be overcome by augmenting with a sufficient number of random samples}
\Figref{fig:pplorderhelpsappendix} shows that while disabling position embedding helps the baseline, similar effects are soon achieved by increasing the random order.
This shows that disabling position embeddings can indeed alleviate some concerns about the order.
This is crucial for pre-trained models, for which position embeddings cannot be ignored.

\begin{figure*}[!htb]
  \includegraphics[width=\linewidth]{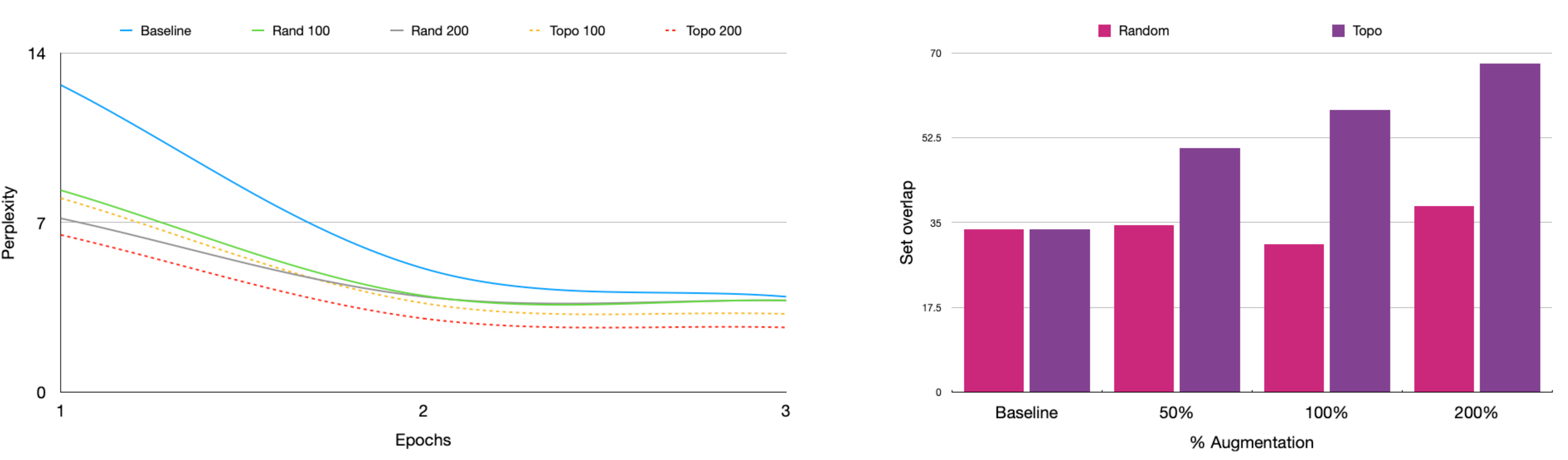}
\caption{Effect of \ours on perplexity and set overlap. \textbf{Left:} Augmentations done \ours helps the model converge faster and to a lower perplexity. \textbf{Right:} Using \ours, the overlap between training and test set increases consistently, while consistently outperforming \rand.}
\label{fig:toposetoverlapandpplappendix}
\end{figure*}

\paragraph{Finding 4: \ours leads to higher set overlap} We next consider blocks of order 2 where the prefix symbol $\ry_p$ is selected randomly as before, but the suffix is set to a special character $\ry_p'$ with 50\% probability.
As the special symbol $\ry_p'$ only occurs with $\ry_p$, there is a high pmi between each $(\ry_p, \ry_p')$ pair as $\pcond{\ry_p}{\ry_p'} = 1$.
Different from finding 1, the output symbols are now shuffled to mimic a realistic setup.
We gradually augment the training data with random and topological orders and evaluate the learning and the final set overlap using training perplexity and Jaccard score, respectively.
The results are shown in \Figref{fig:toposetoverlapandpplappendix}.
Similar trends hold for larger block sizes, and the results are included in the Appendix in the interest of space.

\paragraph{Finding 5: \ours helps across all sampling types} We see from Table~\ref{tab:samplingtypesetoverlapappendix} that our approach is not sensitive to the sampling type used. Across five different sampling types, augmenting with topological orders yields significant gains.

\paragraph{Finding 6: \sts models can learn cardinality and use it for better decoding}

We created sample data from ~\Figref{fig:gmodelappendix} where the length of the output is determined by sum of the inputs $X$.
We experimented with and without including cardinality as the first element.
We found that training with cardinality increases step overlap by over 13\%, from 40.54 to 46.13.
Further, the version with cardinality accurately generated sets which had the same length as the target 70.64\% of the times, as opposed to 27.45\% for the version without cardinality.

\begin{table*}[]
\centering
\begin{tabular}{lrrrrr}
\toprule
         & Beam          & Random        & Greedy        & Top-k        & Nucleus    \\ \midrule
\rand  & $0.39\pm0.05$ & $0.39\pm0.02$ & $0.35\pm0.05$ & $0.39\pm0.02$ & $0.39\pm0.02$ \\
\ours & $0.67\pm0.05$ & $0.67\pm0.05$ & $0.71\pm0.04$ & $0.67\pm0.05$ & $0.68\pm0.05$ \\ \bottomrule
\end{tabular}
\caption{Set overlap for different sampling types with 200\% augmentations. The gains are consistent across sampling types. Similar trends were observed for 100\% augmentation and without positional embeddings. Top-k sampling was introduced by \citep{fan2018hierarchical}, and Nucleus sampling by \citep{holtzman2019curious}.}
\label{tab:samplingtypesetoverlapappendix}
\end{table*} 

\newpage
\clearpage
\section{Fixing the proposal distribution in the \vae formulation}
\label{sec:appendixvae}
\begin{align}
    \log \pcondth{\sY}{\vx}\nonumber &= \log \sum_{\pi_\vz \in \Pi}\pcondth{\pi_{z}(\sY)}{\vx} \nonumber \\
    &= 
    \log  \sum_{\pi_\vz \in \Pi}\frac{q_{\phi}(\pi_\vz)}{q_{\phi}(\pi_\vz)}\pcondth{\pi_{z}(\sY)}{\vx}  \nonumber \\
   &= 
    \log \expval{q_{\phi}(\pi_\vz)}{\frac{\pcondth{\pi_{z}(\sY)}{\vx}}{q_{\phi}(\pi_\vz)}} \nonumber \\
  &\geq  \expval{q_{\phi}(\pi_\vz)}{\log \pcondth{\sY, \pi_\vz}{\vx}} - \expval{q_{\phi}(\pi_\vz)}{\log  q_{\phi}(\pi_\vz)}\nonumber \\
     \log \pcondth{\sY}{\vx}\nonumber &= \log \sum_{\pi_\vz \in \Pi}\pcondth{\pi_{z}(\sY)}{\vx} \nonumber\\ &\geq  \underbrace{\expval{q_{\phi}(\pi_\vz)}{\log\frac{ \pcondth{\pi_\vz(\sY)}{\vx}}{q_{\phi}(\pi_\vz)}}}_{\textsc{elbo}}\nonumber = \Ls(\theta, \phi) \\
\label{eqn:elbo}
\end{align}

Where \eqref{eqn:elbo} is the evidence lower bound~(\textsc{elbo}). The success of this formulation depends on the quality of the proposal distribution $q$ from which the orders are drawn. When $q$ is fixed~(\eg to uniform distribution over the orders), learning only happens for $\theta$. 
This can be clearly seen from splitting \Eqref{eqn:elbo} into terms that involve just $\theta$ and $\phi$:
\begin{align*}
    \nabla_{\phi} \mathcal{L}(\theta, \phi) &= 0 \\
    \nabla_{\theta} \mathcal{L}(\theta, \phi) &= \nabla_{\theta} \expval{q_{\phi}(\pi_\vz)}{\log \pcondth{\sY, \pi_\vz}{\vx}}
\end{align*}

\appendix

\section{Example Appendix}
\label{sec:appendix}

This is an appendix.

\end{document}